\documentclass[a4paper,fleqn]{cas-sc}

\usepackage[numbers]{natbib}

\usepackage{subcaption}
\usepackage{caption}
\usepackage{amsmath}
\DeclareMathOperator*{\argmax}{argmax}
\usepackage{float}
\def\tsc#1{\csdef{#1}{\textsc{\lowercase{#1}}\xspace}}
\tsc{WGM}
\tsc{QE}
\tsc{EP}
\tsc{PMS}
\tsc{BEC}
\tsc{DE}
\begin{document}
\let\WriteBookmarks\relax
\def\floatpagepagefraction{1}
\def\textpagefraction{.001}
\shorttitle{A Generarisable AI-Driven Model for the La Rance Tidal Barrage}
\shortauthors{Túlio Marcondes Moreira et~al}

\title [mode = title]{Development and Validation of an AI-Driven Model for the \\ La Rance Tidal Barrage: A Generalisable Case Study}                      



\author[1]{Túlio Marcondes Moreira}[type=editor,
                        auid=000,bioid=1,
                        prefix=,
                        role=,
                        orcid=0000-0003-4159-1495]
\cormark[1]

\credit{Conceptualisation of this study, Investigation, Methodology, Software, Validation, Writing - Original Draft}

\address[1]{Computer Science Department (DCC), Universidade Federal de Minas Gerais, Belo Horizonte, Minas Gerais, 31270-901, Brazil}

\author[1]{ Jackson Geraldo de Faria Jr}[]

\credit{Investigation, Writing - Review \& Editing}

\author[1]{Pedro O.S. Vaz-de-Melo}[%
   role=,
   suffix=,
   orcid=0000-0002-9749-0151]
\credit{Supervision, Investigation, Writing - Review \& Editing}
\author%
[1]
{ Gilberto Medeiros-Ribeiro}[orcid=0000-0001-5309-2488]

\credit{Supervision, Investigation, Writing - Review \& Editing}


\cortext[cor1]{Corresponding author e-mail address: tmoreira@ufmg.br}


\begin{abstract}
In this work, an AI-Driven (autonomous) model representation of the La Rance tidal barrage was developed using novel parametrisation and Deep Reinforcement Learning (DRL) techniques. Our model results were validated with experimental measurements, yielding the first Tidal Range Structure (TRS) model validated against a constructed tidal barrage and made available to academics. In order to proper model La Rance, parametrisation methodologies were developed for simulating (i) turbines (in pumping and power generation modes), (ii) transition ramp functions (for opening and closing hydraulic structures) and (iii) equivalent lagoon wetted area. Furthermore, an updated DRL method was implemented for optimising the operation of the hydraulic structures that compose La Rance. The achieved objective of this work was to verify the capabilities of an AI-Driven TRS model to appropriately predict (i) turbine power and (ii) lagoon water level variations. In addition, the observed operational strategy and yearly energy output of our AI-Driven model appeared to be comparable with those reported for the La Rance tidal barrage. The outcomes of this work (developed methodologies and DRL implementations) are generalisable and can be applied to other TRS projects. Furthermore, this work provided insights which allow for more realistic simulation of TRS operation, enabled through our AI-Driven model.

\end{abstract}

\begin{graphicalabstract}
\includegraphics[width=\linewidth]{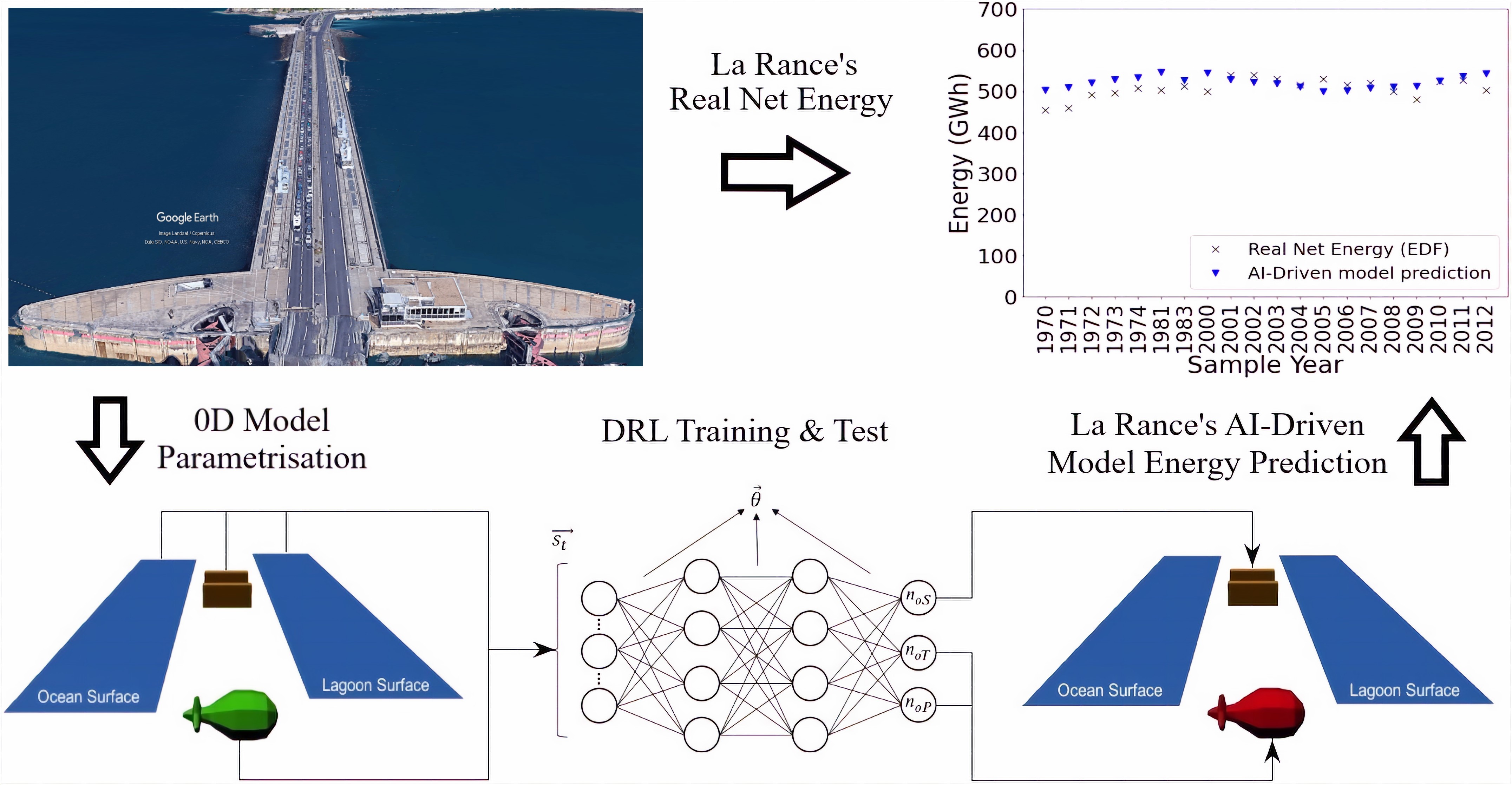}
\end{graphicalabstract}

\begin{highlights}
\item A generalisable AI-Driven model for the La Rance tidal barrage is developed
\item Methodologies for parametrising La Rance's hydraulic structures are introduced
\item A 0D model representation for La Rance is validated against experimental data
\item The AI-Driven La Rance model showcases similar behaviour to the real operation
\end{highlights}


\maketitle

\section{Introduction}

The La Rance tidal barrage is considered one of the greatest achievements of tidal power generation since its debut operation in 1966 \cite{lebarbier1975power, Baker, frau1993tidal, charlier2007forty, rourke2010tidal, neill2018tidal}. Still in full operation today, this Tidal Range Structure (TRS) showcases an installed capacity of 240 $MW$, from its 24 $\times$ 10 $MW$ bulb turbines, generating around 500 $GWh/year$ \cite{EDF500}, at competitive cost of nuclear or offshore wind sources \cite{LaRance20, hendry2016role}. Following the successful operation of La Rance, other TRS projects have also been constructed, such as the 20 $MW$ Annapolis Royal Generating Station and the 254 $MW$ Sihwa Lake Tidal Power Station, in Canada and South Korea, respectively \cite{neill2018tidal, cho2012construction}. Nowadays, La Rance inspires the conceptualisation of other TRS projects around the globe \cite{waters2016tidal, neill2018tidal, li2017ebb}. Among these, the UK stands out with several in depth studies of tidal barrages and lagoons in the following bay and estuaries: Colwyn Bay, Solway, Mersey, Loughor, Duddon, Wyre, Thames \cite{mackie2020potential, waters2016world, aggidis2012tidal, howard2007tidal, aggidis2013operational, sustainable2007turning} and, most notably, the Severn Estuary (e.g. Cardiff, Newport, Fleming and Swansea Bay tidal lagoons, and also the Severn, Cardiff-Weston and Shoots tidal barrages) \cite{angeloudis2017sensitivity, waters2016world,  kelly2012energy, xia2010impact}.

While La Rance is a significant example of how TRS can be economically viable, given enough time of operation, novel projects, such as the Swansea Bay tidal lagoon (SBL), have been questioned if they represent good value for money. This concern is exacerbated by the high initial investment costs that TRS require for construction ($\approx \text{£}1.3$ billion for SBL) and the recent reduction of solar and wind payback periods to $\approx 2 - 20$ (year range) and $\approx13$ years, respectively (considering tax incentives and a maximum lifespan of 30 years) \cite{chang2019evaluation}. These current issues led the UK Government to halt support for the proposed SBL project in 2018 with the argument of high cost of electricity \cite{guardianSBL}. Nevertheless, in a recent turn of events, the Welsh Government launched a tidal lagoon challenge in Wales to ascertain market engagement in the construction of several TRS projects. A total of 55 companies registered for the challenge, with the winner possibly receiving Welsh Government financial support for designing a pathfinder TRS project \cite{civilSBL, walesSBL}. These initiatives demonstrate that TRS development in the near future is still very likely.

For estimating the potential energy of TRS, analytical or numerical models (0D through 3D) have been be considered for simulating such systems. While analytical solutions are useful in estimating an upper bound for the available power  \cite{prandle1984simple, lisboa2017optimal}, numerical models can aid in optimising the operational time-sequence of the hydraulic structures present in a TRS (e.g. turbine and sluices), while accounting for effects not considered in analytical models (e.g. variable bathymetry, pump operational modes). Each of these approaches have different strengths, weaknesses and unique physical simplifications. For instance, while 0D models can be computationally inexpensive, they are solely based on the simple statement of mass conservation for the impounded lagoon. In contrast, 1D through 3D models utilise the shallow water equations to obtain (in an increasing order of detail and computational time) the water profile and velocity components in and out of the lagoon \cite{angeloudis2016numerical, cornett2013assessment, falconer2009severn}. Shallow water models are helpful in assessing environmental impacts of TRS \cite{ma2019impact, ma2019impact2}, while also detecting the sudden high flow-rates developed when turbines start operating – a situation that can abruptly perturb water velocity and elevation (therefore power generation) in the vicinity of hydraulic structures \cite{zhou2014optimization}. Due to the characteristic features of the described approaches, 0D models are usually implemented into optimisation routines of TRS, since these often require a huge number of iterations for convergence \cite{angeloudis2018optimising, harcourt2019utilising, xue2019optimising}. As a next step, 1D to 3D models are employed  to perform verification steps of the operational strategies devised from utilising the 0D model. 2D models are typically picked, among 1D and 3D options, given their satisfactory trade-off between accounting for detailed bathymetry, coastal geography and accurately predicting complex shallow water effects (e.g. funnelling, resonance) \cite{gao2017tidal, cornett2013assessment}, while not being computationally expensive as 3D models.

While the literature has advanced in developing the aforementioned TRS simulation models, these studies have focused on cross verification only (e.g. 0D against 2D) to attest for their validity \cite{angeloudis2016numerical, ahmadian2017optimisation, angeloudis2017comparison, angeloudis2018optimising, xue2019optimising, harcourt2019utilising, xue2020genetic, xue2021design}. This has been the case, since experimental data from constructed TRS are scarce and under the intellectual protection of few companies around the globe. To fulfil this gap in the literature, we develop in this paper an artificial intelligence (AI) driven model representation of the La Rance tidal barrage, utilising an updated Deep Reinforcement Learning (DRL) approach \cite{moreira2021prediction} and experimental data from the literature \cite{lebarbier1975power, swane2007tidal, bosc1997groupes, rolandez2014discharge}. The development of such model occurs in two stages. First, a series of generalisable methodologies for simulating turbines (in power generation and pumping modes), transition ramp functions (for opening and closing the TRS hydraulic structures) and equivalent lagoon wetted area are developed and assembled into a parametrised 0D model of the La Rance tidal barrage. The 0D La Rance model is then validated in its capabilities of predicting (i) power output and (ii) lagoon water level variations, using measurements from \cite{lebarbier1975power} as reference. In a second stage, a novel DRL implementation for optimising the operation of La Rance's hydraulic structures utilising the Unity ML-Agents package, based on \cite{moreira2021prediction}, is presented and discussed. The operational strategy learned by the DRL agent for controlling the sequence and timing of operation of the hydraulic structures (i.e. the AI-Driven La Rance model) is shown to be closer to the real strategy utilised in La Rance, in comparison over state-of-art approaches proposed by the literature for other TRS \cite{xue2019optimising, angeloudis2018optimising}. Furthermore, annual energy predictions from the AI-Driven model are also compared against measured data from La Rance, showcasing good agreement of results.

\section{Deep Reinforcement Learning}\label{DRLback}

In recent years, DRL techniques have represented a breakthrough for AI methods, performing optimally in several real-time problems that previously could only be solved by experienced human operators. The potential of DRL techniques for solving such tasks has received a significant attention in applications with games, such as Atari, Chess, Shogi, Game of Go and StarCraft II \cite{mnih2013playing, silver2016mastering, silver2018general}, where smart DRL agents were able to showcase a performance superior to human players. Amid the advancements of DRL algorithms, complex real-life optimisation problems also started to be tackled, such as trading and finance, self-driving cars, healthcare and energy efficiency (e.g. heating, ventilation, air-conditioning and datacentre cooling systems) \cite{deng2016deep, sallab2017deep, esteva2019guide, zhang2019whole, li2019transforming}. In the context of renewable energy, reinforcement learning has also been used for optimising the operation of smart grids, wind turbines, solar panels, stream turbines and more recently, TRS \cite{lu2018dynamic, saenz2019artificial, shresthamali2017adaptive, phan2020deep, nambiar2017reinforcement, moreira2021prediction}.

\begin{figure}[h]
	\centering
	\includegraphics[width=.5\linewidth]{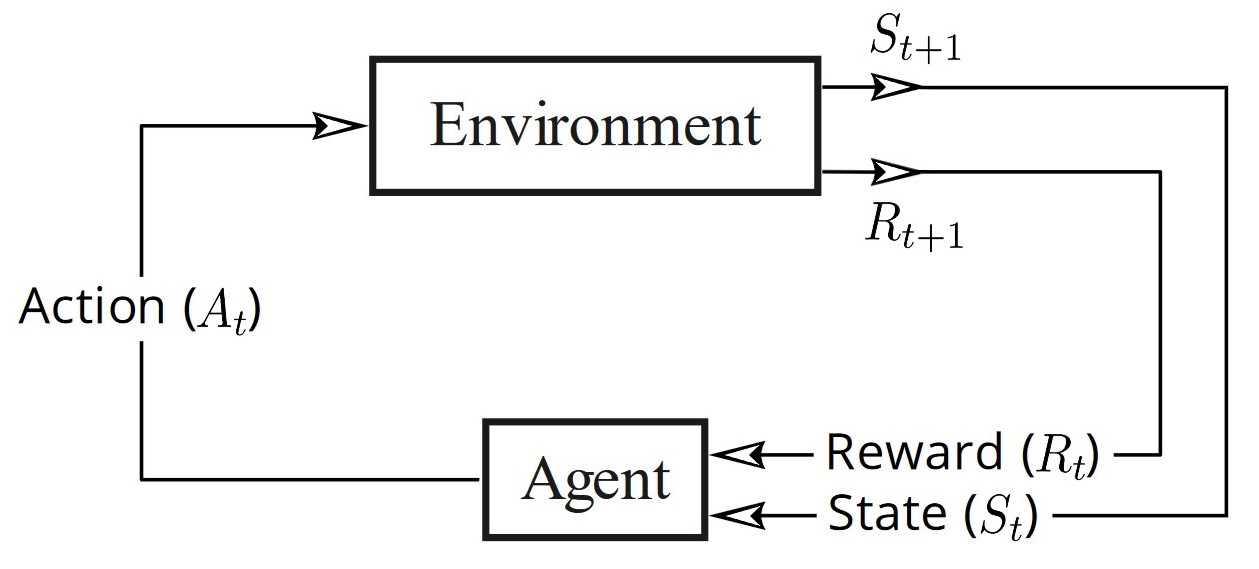}
	\caption{General agent-environment interaction in an MDP, showcasing the resulting state-action, next reward triples sequence. Edited from \cite{sutton2018reinforcement}.}
	\label{RLFramework} 
\end{figure}

In general, in order to assess if an operational optimisation problem can be solved through DRL, the former needs to be mapped into a Markov Decision Process (MDP). As shown in \cite{sutton2018reinforcement}, the mathematical formalisation of an MDP is composed of (i) an agent capable of actions ($A_t$) and (ii) a reactive environment, that outputs new environmental states ($S_{t+1}$) and rewards ($R_{t+1}$) based on the received action $A_t$ from the agent. The defined quantities $A_t$, $S_t$ and $R_t$ are random variables, with well-defined probability distributions. The interaction between agent and environment can be visualised in a general agent-environment MDP framework in Fig. \ref{RLFramework}.

As is the case with this work, the simulation of the environment (for acquiring $S_{t+1}$ and $R_{t+1}$) can be performed with numerical models \cite{mnih2013playing, silver2016mastering, silver2018general}. The employment of numerical models allow for utilising model-free DRL methods, where the MDP's agent learns by experience -- sampling states and rewards from the environment for every action $A_t$. In non-deterministic scenarios, the selection of possible actions by the agent (as a function of input states) is a stochastic process known as policy, mathematically described as a conditional probability distribution of the form:
\begin{equation}
\pi(a_t|s_t).
\label{Policy}
\end{equation}

With a defined (i) policy (for choosing the agent's actions) and (ii) a numerical model of the environment (model-free DRL approach), simulation sequences of the MDP can be run (Fig. \ref{RLFramework}). At each cycle of the MDP, multiple time-steps $t = 0, 1, 2, 3 ...$ observations ($O_i$) of the agent-environment interaction can be sampled. These are organised as a sequence of state-action, next reward triples:
\begin{equation}
O_i = < s_i, a_i, r_{i+1} >,
\label{Observation}
\end{equation}
where, $s_i, a_i, r_{i+1}$ are instances of the random variables ($S_t$, $A_t$ and $R_{t+1}$). While the sequence of state-action pairs defines a trajectory $\tau$:
\begin{equation}
\tau = s_0, a_0, s_1, a_1, s_2, a_2 ...,
\label{Trajectory}
\end{equation}
the summation of the observed sequence of rewards yield the total return $G_t$ at time-step $t$:
\begin{equation}
G_t = R_{t+1} + \gamma R_{t+2} + \gamma^2 R_{t+3} ... = \sum^\infty_{k=0} \gamma^k R_{t+k+1},
\label{Return}
\end{equation}
where $\gamma$ is a discount factor between 0 and 1.

The objective of reinforcement learning problems is to find an optimal policy $\pi^*$ that maximises the expected return of rewards $E[G_t]$ conditioned on any initial state, i.e.
\begin{equation}
\pi^* = \argmax_{\pi} E_{\pi} [G_t | S_t = s_t], \forall s_t.
\label{GoalPolicy}
\end{equation}

A policy $\pi^*$ for the agent can be obtained (i) indirectly, through state-value or action-value functions (e.g. Deep Q-Network) \cite{mnih2013playing} or (ii) directly, with policy based methods (e.g Policy Gradient) \cite{sutton2000policy}. In this work, Proximal Policy Optimisation (PPO) \cite{schulman2017proximal} (a Policy Gradient algorithm) is utilised for training the DRL agent. In PPO, a neural network representation (actor neural network) is utilised for parametrising the policy, enabling the sampling actions of $A_t$ as a function of environmental inputs $S_{t}$. For the case of discrete actions (as is the case in this work), the neural network outputs the probabilities of each possible action in that state using a softmax layer. A full mathematical description of the PPO algorithm, in the context of TRS application and for both training and test stages, can be found in \cite{moreira2021prediction}. The agent-environment MDP modelling and training stages for the La Rance TRS are described in Section \ref{MDP}.

\section{State of Art TRS Simulation and Operation.}\label{StateTrs}

In this section, state of art methods for simulating, operating and parametrising the hydraulic structures that compose a TRS are detailed and discussed. While some of the methods are adopted in this work (e.g. 0D TRS model), others are contrasted against experimental measurements from the La Rance tidal barrage, indicating the need of developing novel approaches that are capable of yielding more realistic results.


\subsection{0D Simulation}\label{0DModel}
Derived from mass conservation, 0D models correspond to the simplest numerical model representation for TRS, capable of estimating lagoon water motion as a function of total flow rate from turbines and sluices. Since turbine models predict flow-rate and power as a function of water head between ocean and lagoon, 0D models also allow for estimating power production of TRS subjected to water head variations. When the goal is the optimisation of TRS operation for maximising power generation or revenue, 0D models are usually chosen, given their computational efficiency. Furthermore, 0D models have presented good agreement of results with more complex finite-element 2D models when considering ``small-scale'' projects (e.g. Swansea Bay Tidal Lagoon)  \cite{angeloudis2017comparison, angeloudis2018optimising, xue2019optimising, neill2018tidal, andrea2019implementation}. Formally, 0D models can be written as:
\begin{equation} 
  \frac{d L}{d t} = \frac{Q_T}{Al(L)},
  \label{0D}
\end{equation}
where $L$ is the lagoon water level (in meters), $Q_T$ is the total directional water flow rate ($m^3/s$), predicted for both turbines and sluices using function approximations, and $Al(L)$ is the variable lagoon area ($m^2$). From Eq.~(\ref{0D}), the lagoon water level at the next time-step ($L_{t+1}$) can be numerically calculated by a backward finite difference method:
\begin{equation} 
  L_{t+1} = L_t + \frac{Q_T}{Al(L)} \Delta t,
  \label{Num0D}
\end{equation}
where $L_{t}$ is the water level at time-step $t$ and $\Delta t$ the discretized time. In this work, Eq.~(\ref{Num0D}) represents the simulation environment for our MDP.

\subsection{Two-Way with Pumping Operation Scheme}

\begin{figure}[]
  \centering
  \begin{subfigure}[t]{.49\linewidth}
    \centering\includegraphics[width=\linewidth]{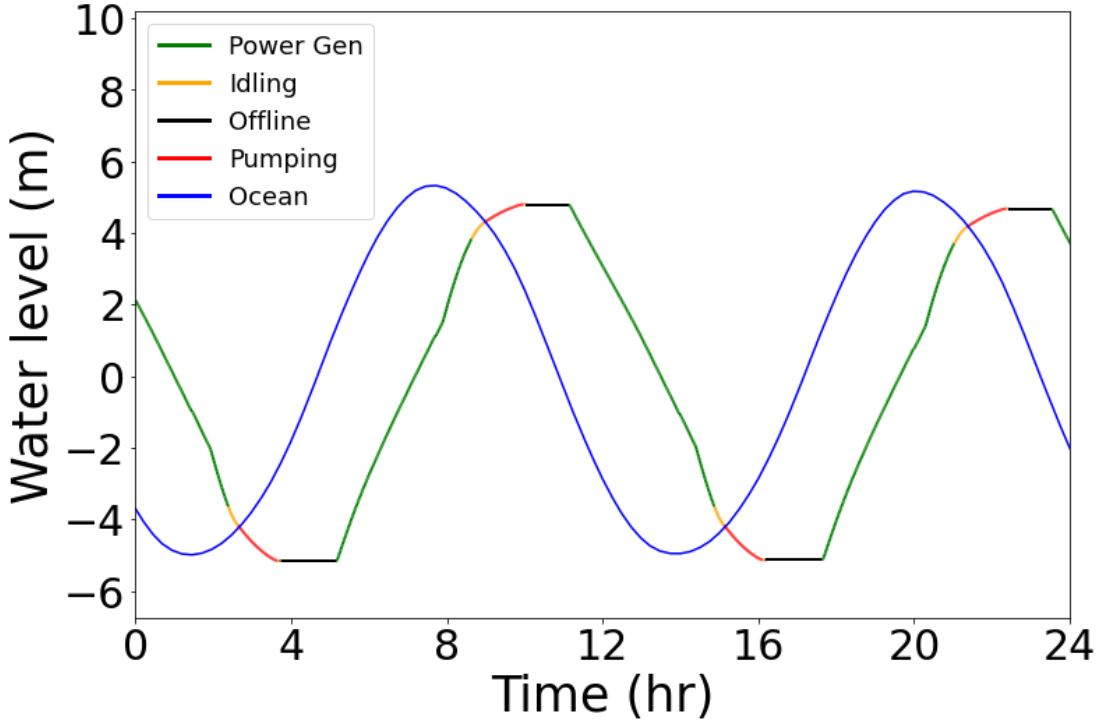}
    \caption{Coloured lagoon water levels following turbine operation.} \label{OpTurbine}
  \end{subfigure} \hfill \bigskip
  \begin{subfigure}[t]{.49\linewidth}
    \centering\includegraphics[width=\linewidth]{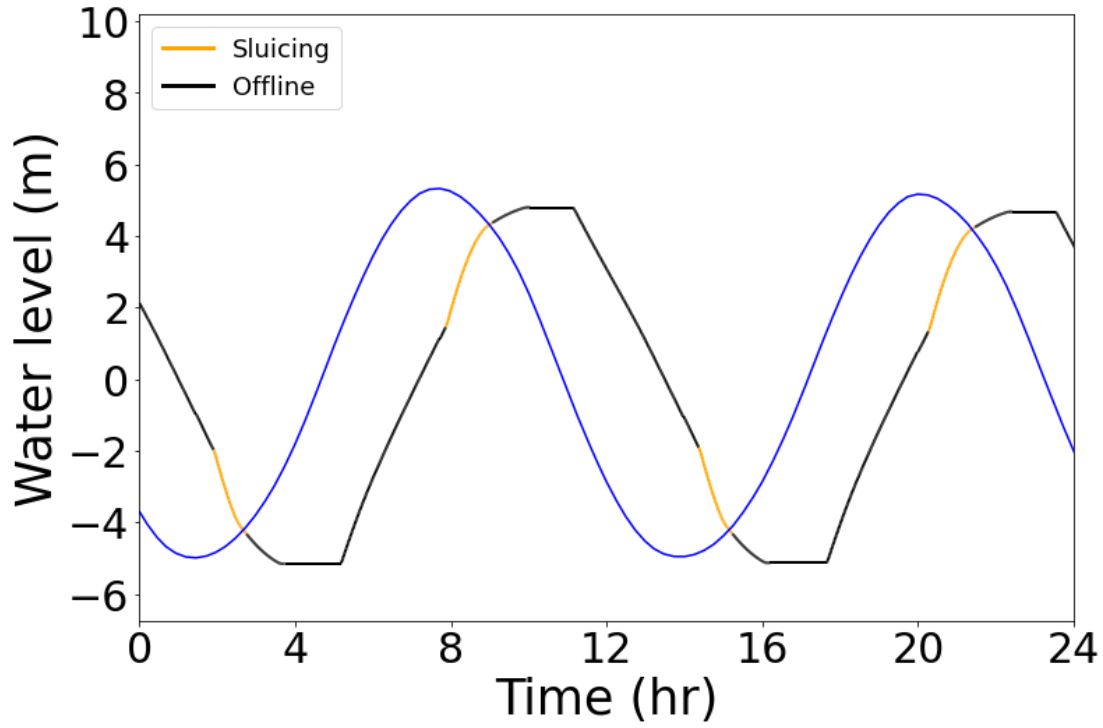}
    \caption{Coloured lagoon water levels following sluice operation.}\label{OpSluice}
  \end{subfigure}
  \caption{General $T.W.P$ scheme operation of TRS, following state-of-art constraints. Ocean is coloured in blue, while lagoon water levels are shown and coloured following turbines and sluices operational modes, with green representing power generation mode, orange -- idling/sluicing mode, black -- offline mode and red -- pumping mode.} \label{OpStateOfArt}
\end{figure}

Given a 0D TRS model, state-of-art optimisation routines require a fixed TRS scheme of operation as a basis for optimising the operational time sequence of turbines and sluices. From the literature \cite{angeloudis2018optimising, xue2019optimising, harcourt2019utilising}, two-way with pumping schemes ($T.W.P$) have showcased the best results for maximising either power or revenue generation. $T.W.P$ schemes are characterised by the ability of the TRS to generate power either during the flood or ebb tides, with the possibility of utilising turbines as pumps to increase the water head difference between ocean and impounded lagoon. As discussed in the seminal work of Gibrat from 1955 \cite{gibrat1955tidal} (prime chief investigator of La Rance Tidal Barrage project), the augmentation of TRS operation with pumping capabilities can increase energy extraction of these systems significantly. Indeed, state-of-art research \cite{angeloudis2018optimising, xue2019optimising} have supported this claim, simulating that a $20-40\%$ increase of energy output is expected when implementing pumping in classic two-way ($T.W$) scheme approaches. However, the literature interpretation of $T.W.P$ schemes is still not aligned with observed measurements of the only case study in the world: La Rance. For instance, in Figs. \ref{OpTurbine} and \ref{OpSluice}, predicted lagoon water level variations for the state-of-art literature interpretation of $T.W.P$ schemes are coloured according to the operational mode chosen for turbines and sluices, respectively, with the sequence of operations for the hydraulic structures being dictated by literature constraints \cite{angeloudis2018optimising, xue2019optimising}. While the logic of operation for turbines and sluices is mostly identical to the observed in La Rance (green, orange and black regimes in Figs. \ref{OpTurbine} and \ref{OpSluice}), literature constraints limit turbine pumping stage to occur only when negative head differences (against gravity) are observed (red regimes in Fig. \ref{OpTurbine}). In contrast, measurements from the La Rance TRS \cite{lebarbier1975power, bosc1997groupes}, showcase pump operation in both positive and negative water head scenarios, with pump shutoff negative heads (null pump flow rate) up to $6m$. Furthermore, state-of-art optimisation routines need to choose to fix either power input or pump flow-rate to reduce computational costs \cite{angeloudis2017comparison} -- differently from what is observed in La Rance \cite{lebarbier1975power}, where fine tuned power input is provided for turbines in pump mode. In Section \ref{LaRanceDRL}, we show that the operational strategy devised by our trained DRL agent is capable of filling this gap in the literature, operating the 0D La Rance model in a more realistic fashion.

\subsection{Hydraulic Structures Parametrisation}
With a given TRS model and operational scheme, function approximations for simulating turbines and sluices are required. Regarding TRS turbines in pump mode, literature research has been limited. In fact, recent studies have simulated pump operation either by fixing the efficiency (for any head difference) to $70\% - 40\%$ \cite{angeloudis2017comparison, yates2013energy}, or by directly following experimental pump efficiency curves from \cite{yates2013energy}. For negative head differences, the literature has also adopted an idealised pump efficiency interpretation \cite{angeloudis2018optimising} (disregarding efficiency variations with pump rotation), cast as:
\begin{equation} 
  \eta_p = \frac{P_{out}}{P_{in}},
  \label{PumpEff}
\end{equation}
where $P_{in}$ is the electrical power available to the pump and $P_{out}$ is the rate of work exerted from the pump to the fluid:
\begin{equation} 
  P_{out} = \rho g Q_p |h_p|,
  \label{PumpOutput}
\end{equation}
where $\rho$ is the seawater density ($1024~kg/m^3$), $g$ the gravity acceleration ($9.81~m/s^2$), $Q_p$ the pump flow rate and $h_p$ the negative head surpassed during pumping. From Equations\,\ref{PumpEff} and \ref{PumpOutput} we obtain:
\begin{equation} 
  Q_p = \frac{\eta_p P_{in}}{\rho g |h_p|}.
  \label{PumpFlow}
\end{equation}

Although Eq.~\ref{PumpFlow} can help estimate pump flow rate, given a negative head and input power, it cannot explain maximum pump shutoff heads (where $Q_p = 0$), or maximum pump flow rates that occur when $h_p = 0$. These issues, combined with the fact that turbine specifications (e.g. diameter, capacity, applied power input) were omitted in the \cite{yates2013energy} study, and that pump efficiencies are only available for negative heads in the small range [$\approx .3m, \approx 1.9m$], indicate the urgent need of more accurate and generalisable TRS pump models. In order to fill this gap, a novel and generalisable TRS pump model is developed in Section \ref{PLaRance} and applied to our parametrised 0D La Rance model.

Besides ``pumping mode'', turbine modelling also require functions for simulating ``power generation'' and ``idling'' modes of operation. In the literature, turbine hill charts (e.g Sulzer Escher Wyss of Zurich, Andritz \cite{Baker, aggidis2012tidal}) have been used for estimating power output of bulb turbines of various diameters. In Section \ref{TLaRance}, parametrisation techniques based on experimental measurements from La Rance are utilised for deriving hill charts for the La Rance tidal barrage. When operating in ``idling'' mode, turbines act as sluices, aiding the sluicing stage of operation. For estimating flow rates in this mode, the orifice equation has commonly been employed:
\begin{equation}
Q_o = C_dA_S\sqrt{2g|h|},
\label{Orif}
\end{equation}
where flow rate $Q_o$ is a function of $h$ (the head difference between ocean and lagoon), $C_d$ is the dimensionless discharge coefficient (greatly dependent on sluice gate design \cite{Baker}), and $A_S$ the sluice/orifice area. The choice of best $C_d$ values for turbines and sluices for the La Rance case study are presented in Section \ref{SLaRance}. Finally, in order to more accurately simulate the starting and end of operation of the TRS hydraulic structures, a novel momentum ramp function is developed and presented in Section \ref{MomRampS}.

\section{La Rance Tidal Barrage Components Parametrisation} \label{ParamDef}

In this section, we propose techniques to reverse engineer the hydraulic structures that compose a real TRS into parametric functions. The available data utilised in this section is mostly obtained from a 1975 study \cite{lebarbier1975power}, executed in collaboration with Electricité de France (EDF), the company responsible for La Rance operation still to this day. Data from this study are utilised in order to create parametric models for turbines in ebb and flood operation (in both power generation and pump modes), sluices and equivalent impounded wetted area. Furthermore, a novel ramp function, named momentum ramp function, is developed in order to help estimate flow rate and power variations when opening/closing TRS's hydraulic structures.

\begin{figure}[h]
  \centering
  \begin{subfigure}[t]{.49\linewidth}
    \centering\includegraphics[width=\linewidth]{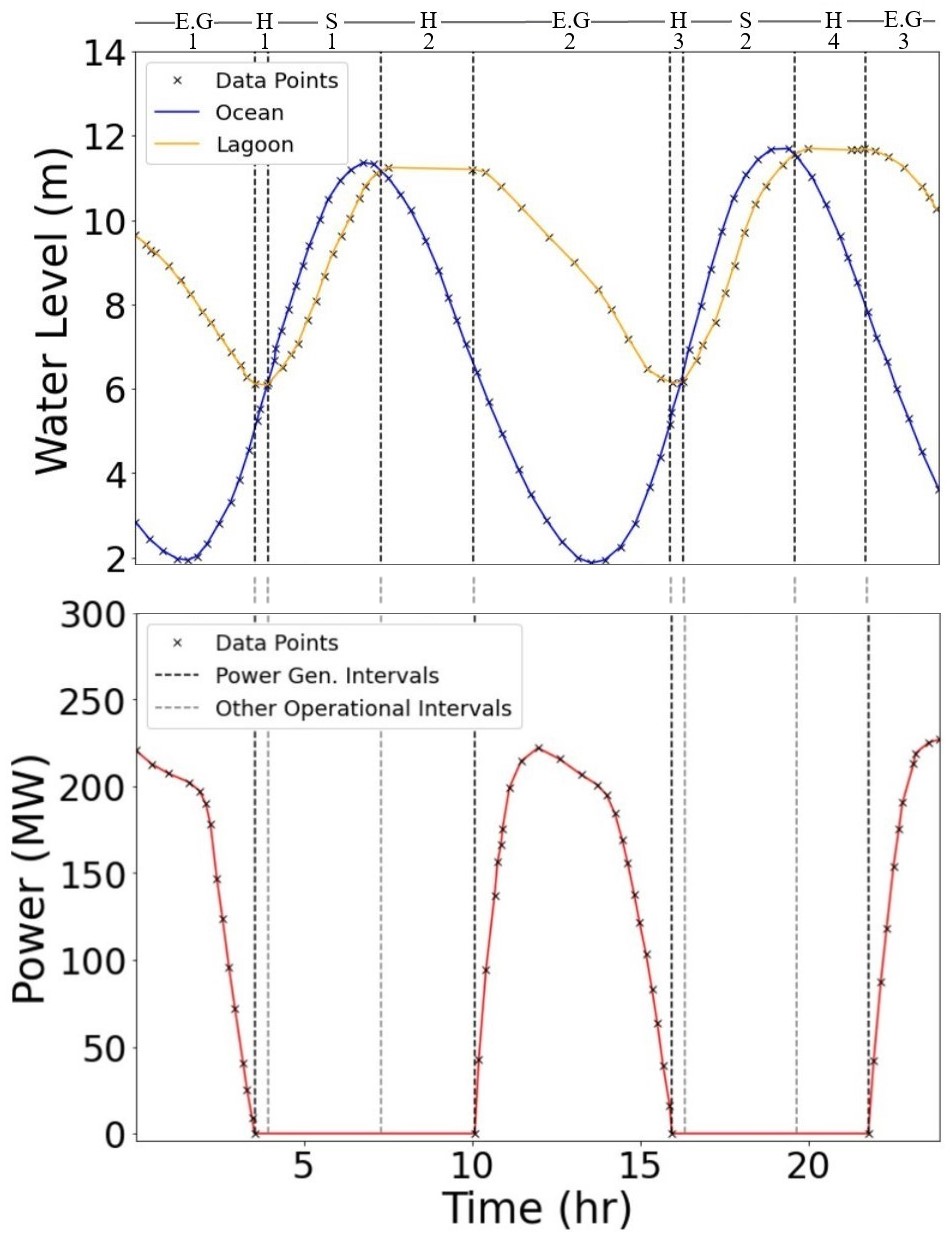}
    \caption{Ocean and Lagoon Water Levels and corresponding Power generation for $E_oG$ scheme.} \label{EGRwl}
  \end{subfigure}  \hfill
  \begin{subfigure}[t]{.49\linewidth}
    \centering\includegraphics[width=\linewidth]{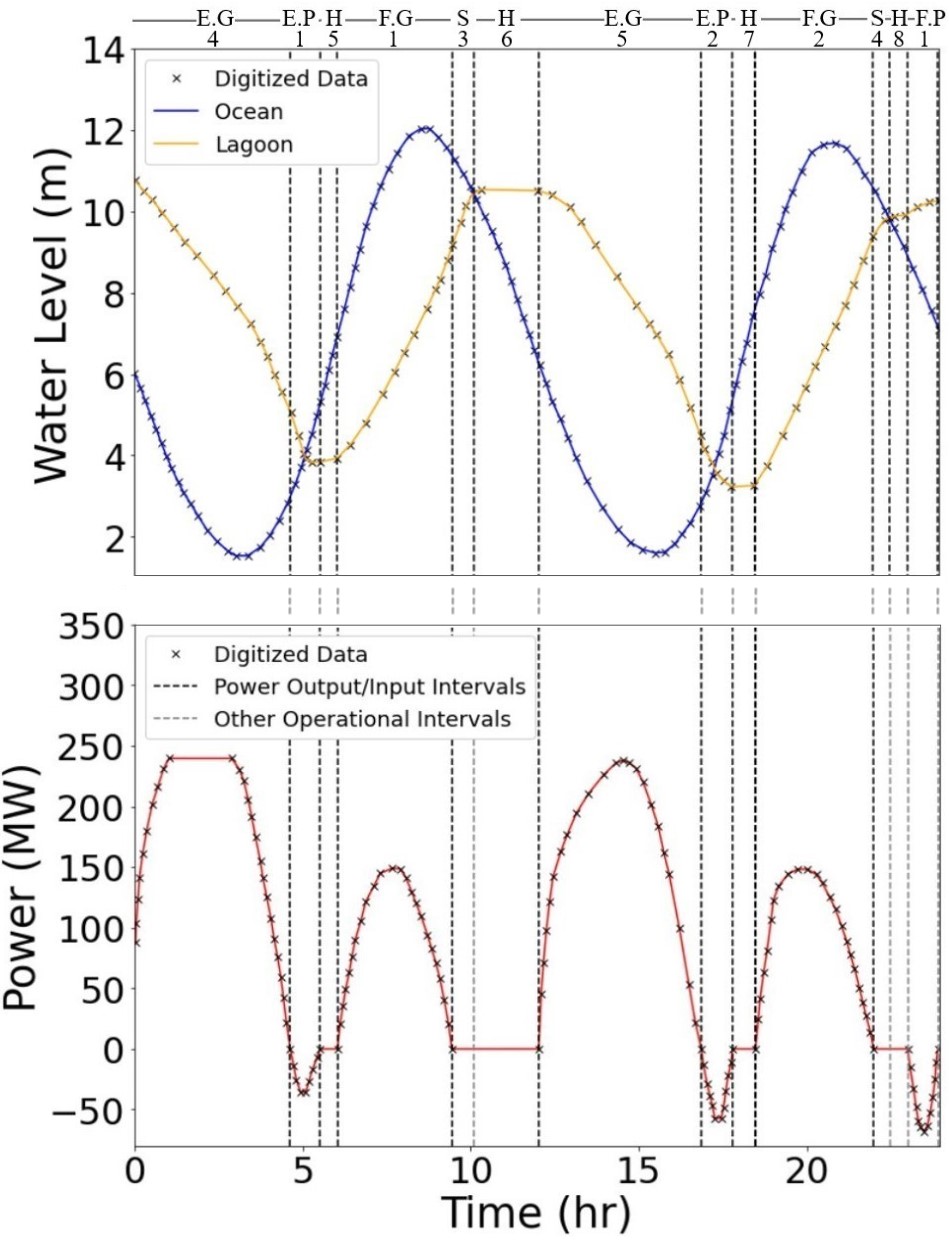}
    \caption{Ocean and Lagoon Water Levels and corresponding Power generation for $T.W.P$ scheme.}
    \label{EGRPow}
  \end{subfigure} \bigskip
  \caption{Measurements of $E_oG$ and $T.W.P$ schemes of operation at La Rance. Digitised from \cite{lebarbier1975power}.} \label{TWPRance}
\end{figure}

From \cite{lebarbier1975power}, the observed variations of lagoon water level and power output/input for two days of observations are digitised, yielding Figs. \ref{EGRwl} and \ref{EGRPow}. In Fig. \ref{EGRwl}, the observed operational modes correspond to a conventional ebb-only generation ($E_oG$) scheme, while Fig. \ref{EGRPow}  showcase a $T.W.P$ scheme. The resolution of the digitised data is represented by a ``$\times$'' label. The initials for the operational modes that were set for each scheme of operation are also presented at the top of each image. The detailings of each labelled initial, numbered in order of occurrence in Figs. \ref{EGRwl} and \ref{EGRPow}, are shown in Table \ref{OperatingUnits}.

\begin{table}[width=.6\linewidth,cols=2,pos=h]
\caption{Operational Modes for $E_oG$ and $T.W.P$ La Rance's Schemes of Operation.}\label{OperatingUnits}
\begin{tabular}{cccc}
\hline
\multirow{2}{*}{Label} & \multirow{2}{*}{Operational Mode} & \multicolumn{2}{c}{Operating Hydraulic Units$^{a}$} \\ \cline{3-4} 
                       &                          & \textbf{$E_oG$}          & \textbf{$T.W.P$}          \\ \hline
$E.G$                    & Ebb Generation           & $T$: ON \& $S$: OFF     & $T$: ON \& $S$: OFF $\rightarrow$ ON  \\
$F.G$                    & Flood Generation         & $N.A.$                     & $T$: ON \& $S$: OFF $\rightarrow$ ON  \\
$H$                      & Holding                  & $T$: OFF \& $S$: OFF    & $T$: OFF \& $S$: OFF                     \\
$S$                      & Sluicing                 & $T$: ON \& $S$: ON  & $T$: ON \& $S$: ON  \\
$E.P$                    & Ebb-Oriented Pumping     & $N.A.$  & $T$: ON \& $S$: ON $\rightarrow$ OFF  \\
$F.P$                    & Flood-Oriented Pumping   & $N.A.$                     & $T$: ON \& $S$: ON $\rightarrow$ OFF  \\ \hline
\multicolumn{4}{l}{$^{a}$ For each scheme of operation \cite{Baker, lebarbier1975power}.} \\
\multicolumn{4}{l}{$T$: Turbines. $S$: Sluices. $N.A.$: Not Applicable.}
\end{tabular}
\end{table}

\begin{table}[width=.3\linewidth,cols=2,pos=h]
\caption{La Rance Barrage Design.}\label{LaRanceDesign}
  \begin{tabular*}{\tblwidth}{@{} LLLL@{} }
  \toprule
	$N^o$ of Turbines & $24$  \\
	Turbine speed ($rpm$) & $94$  \\
	Turbine Diameter ($m$) & $5.35$  \\
	Turbine Capacity (MW) & $10$ \\
	Max. Pump Head ($m$) & $6$ \\
	Sluice Area ($m^2$) & $900$  \\
  \bottomrule
\end{tabular*}
\end{table}

It is worth emphasising that during $E.G$ at the $E_oG$ scheme, turbines are expected to operate alone. This contrasts with the $T.W.P$ scheme, where at the end of $E.G$ and $F.G$, turbines are expected to operate together with sluices (i.e., the variant operation of TRS described in \cite{moreira2021prediction}) \cite{Baker, lebarbier1975power}.

\begin{figure}
  \centering
  \begin{subfigure}[t]{.49\linewidth}
    \centering\includegraphics[width=\linewidth]{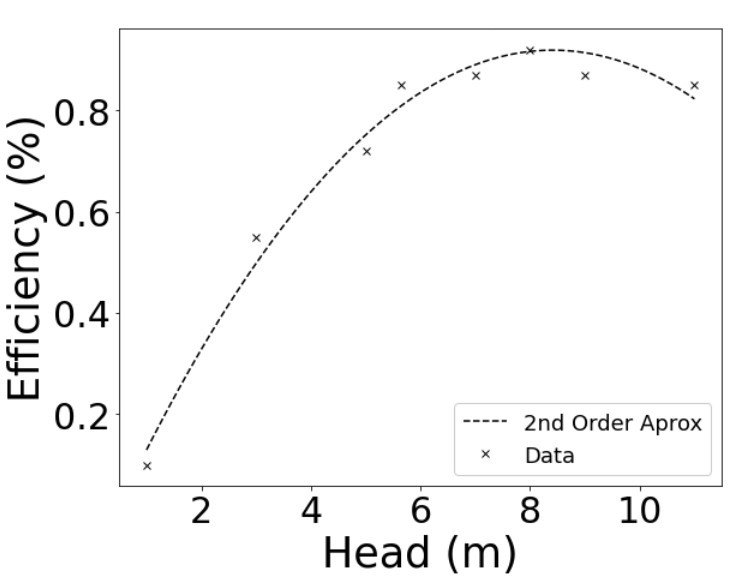}
\caption{$E.G$: $Eff_{E.G} = -.0144 h^2 + .2417 h + .0981$} \label{EGEff}
  \end{subfigure} \hfill \bigskip
  \begin{subfigure}[t]{.49\linewidth}
    \centering\includegraphics[width=\linewidth]{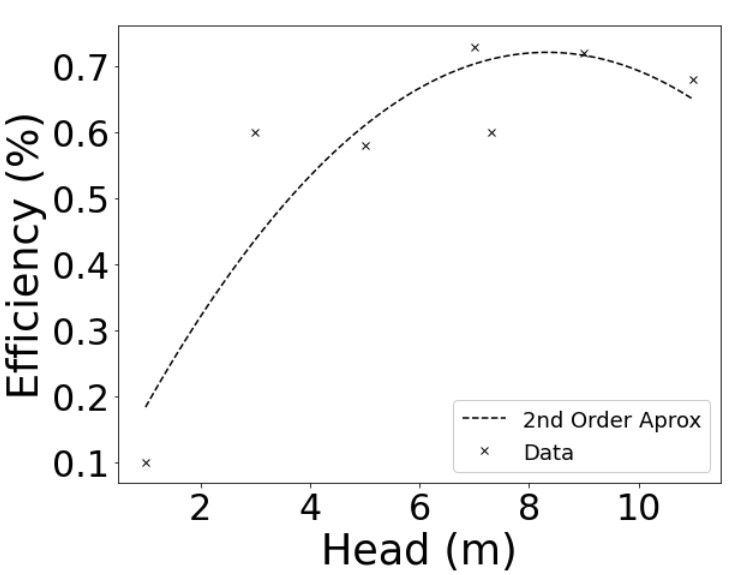}
    \caption{$F.G$: $Eff_{F.G} = -.01 h^2 + .167 h + .0259$}
    \label{FGEff}
  \end{subfigure}
  \caption{$E.G$ and $F.G$ turbine efficiency ($E_{ff}$) for La Rance, from \cite{lebarbier1975power, swane2007tidal}.} \label{EffRance}
\end{figure}

\begin{figure}
	\centering
	\includegraphics[width=.5\linewidth]{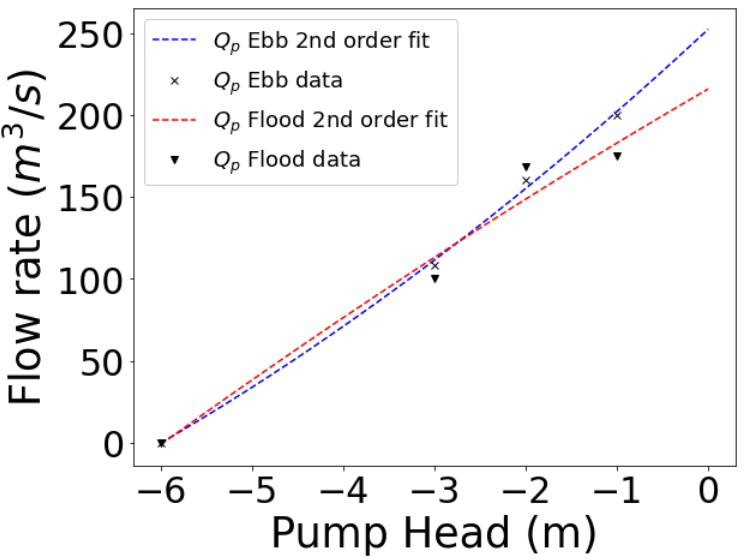} \bigskip
	\caption{$E.P$ and $F.P$ turbine flow rate estimates as a function of water head. ($Q_{pEbb} = 1.6 h_p^2 + 51.8h_p + 252.2$) and ($Q_{pFlood} = -0.6 h_p^2 + 32.4h_p + 215.8$).}
	\label{EbbPumpFlow}
\end{figure}

Design specifications for La Rance, taken from \cite{lebarbier1975power}, are shown in Table \ref{LaRanceDesign}. Also, \cite{lebarbier1975power, swane2007tidal} present data showcasing turbine efficiency as a function of water head $h$ (for $E.G$ and $F.G$ modes of operation). From these data, $2nd$ order approximations are derived and shown in Figs. \ref{EGEff} and \ref{FGEff}. Furthermore, data presenting the expected pump flow rates $Q_p$ as a function of $h_p$ for La Rance's bulb turbine in $E.P$ and $F.P$ operational modes, with fixed (maximum) power input of $P_{in} = 6~MW$, are also available in Table \ref{MeasuredPump}. From these data, $2nd$ order approximations are derived and shown in Fig. \ref{EbbPumpFlow}. Following Table \ref{MeasuredPump} and Fig. \ref{EffRance}, we note that pump shutoff head is $h_s = -6~m$ for both $E.P$ and $F.P$ modes, and that maximum flow rates, when $h_p = 0~m$, measure $Q_{M}Ebb = 252.2 ~m^3/s$ and  $Q_{M}Flood = 215.8 ~m^3/s$, respectively.

\begin{table}[width=.4\linewidth,cols=5,pos=h]
\centering
      \caption{Measured $E.P$ and $F.P$ flow rates at La Rance, for a fixed power input $P_{in} = 6MW$ (from \cite{lebarbier1975power, bosc1997groupes}).}\label{MeasuredPump}
\begin{tabular}{c|cccc}
\hline
$Q_p (h_p, P_{in} = 6MW)$              & \multicolumn{4}{c}{Against $h_p (m) < 0$} \\
 & -6          & -3           & -2           & -1           \\ \hline
$Q_p Ebb~(m^3/s)$      & 0           & 108          & 160          & 200          \\
$Q_p Flood~(m^3/s)$             & 0           & 100          & 168          & 175          \\ \hline
\end{tabular}
\end{table}

\begin{table}[width=.4\linewidth,cols=6,pos=h]
\centering
  \caption{Stored volume of water for the La Rance estuary, from \cite{lebarbier1975power}.}\label{TidalPrism}
\begin{tabular}{l|lllll}
\hline
$\Delta V_i (10^6 m^3)$ & 0 & 65 & 110 & 150  & 184  \\ \hline
$z_i (m)$ & 0 & 5  & 8.5 & 10.9 & 13.5 \\ \hline
\end{tabular}
\end{table}

\cite{lebarbier1975power} also provides various estimates for the volume ($V_i$) of water stored in the La 
Rance estuary as ocean tide fluctuates to a height $z$, defined from the lowest tidal level in the equinoctial low tide ($z = 0$), shown in Table \ref{TidalPrism}. 

Finally, an upper bound for the maximum turbine flow rate (for any mode of operation) is set to $280~m^3/s$, utilising site measurements from La Rance as reference \cite{rolandez2014discharge}. In the following sections, the data presented will be utilised to parametrise all elements that compose the 0D La Rance model. As a first step, since sluices and turbines (in both power generation and pump modes) require a ramp function to be simulated, a novel momentum ramp function for the hydraulic structures is presented in the next section.

\subsection{Momentum Ramp Function}\label{MomRampS}

In order to model flow rate variations from opening/closing the hydraulic structures that compose a TRS, a ramp function is required. Given that solutions from the literature try to solve this problem with a heuristic approach \cite{zhou2014refinements, angeloudis2018optimising, xue2019optimising}, we propose a new derivation for a ramp function more grounded on physical principles. In order to do so, we utilise an electro-hydraulic analogy of a direct current $LR$ circuit, where the circuit's inductance ($L$) and resistance ($R$) are analogous to the hydraulic inertance and resistance, respectively. These types of analogies have been extensively used in the literature, with applications for both pipe fluid flow and open channel flow \cite{rodda1987water,rodriguez1979analogy}. The derived ramp function is an expansion of the heuristic momentum ramp function presented in \cite{moreira2021prediction}. A complete derivation of the proposed ramp function is presented in the Supplementary Material. 

The obtained numerical form of the momentum ramp function is presented as:
\begin{equation}\label{MomRamp}
   Q (t+1) = Q_{ss} + \left(Q (t) - Q_{ss} \right)e^{-\Delta t/\zeta},
\end{equation}
where $Q (t+1)$ represents the flow rate at the next time-step, $Q (t)$ the flow rate at the present time and $\zeta$ controls the intensity of flow-rate updates. The term $Q_{ss}$ represents the steady-state flow rate estimate for the hydraulic structure as a function of head difference. As an example: for sluices, $Q_{ss}$ can be obtained from the orifice equation (Eq.~\ref{Orif}), while for turbines in power generation mode, $Q_{ss}$ can be obtained from parametrised turbine hill charts. Furthermore, since in steady-state regime power output relates linearly to turbine flow rate, the numerical form of the momentum ramp function can also be applied to predict the power output evolution:
\begin{equation}\label{MomRampPow}
   P (t+1) = P_{ss} + \left(P (t) - P_{ss} \right)e^{-\Delta t/\zeta},
\end{equation}
where $P (t+1)$ is the power output at the next time-step and $P (t)$ the power output at the present time. The term $P_{ss}$ represents the steady-state power output estimate for the turbine as a function of head, i.e. a turbine hill chart estimate.

The numerical form of the momentum ramp function in Eq.~(\ref{MomRamp}) simplifies the very complex phenomena of opening/closing the turbines and sluices that compose a TRS. Indeed, beyond the hydraulic resistance and inertance of these systems, the opening/closing of turbines and sluices also involve adjusting the pitching of runner blades/guide vanes (for turbines) and aperture of gates (for sluices). Nevertheless, we show in following sections that the developed ramp function can accurately help estimate power output and flow rate evolution for both starting/closing stages of turbines and sluices, given appropriate $\zeta$ values. As a lower bound for $\zeta$ estimates (for both sluices and turbines), we utilise the value $\zeta = 1.091~min$, which guarantees a precision of $10^6$ for the complete opening/closing of hydraulic structures in a $15~min$ time interval (analogous to the tuned ramp function in \cite{moreira2021prediction}).

\subsection{Turbines -- Power Generation Mode}\label{TLaRance}

\begin{figure}[h!]
  \centering
  \begin{subfigure}[t]{.49\linewidth}
    \centering\includegraphics[width=\linewidth]{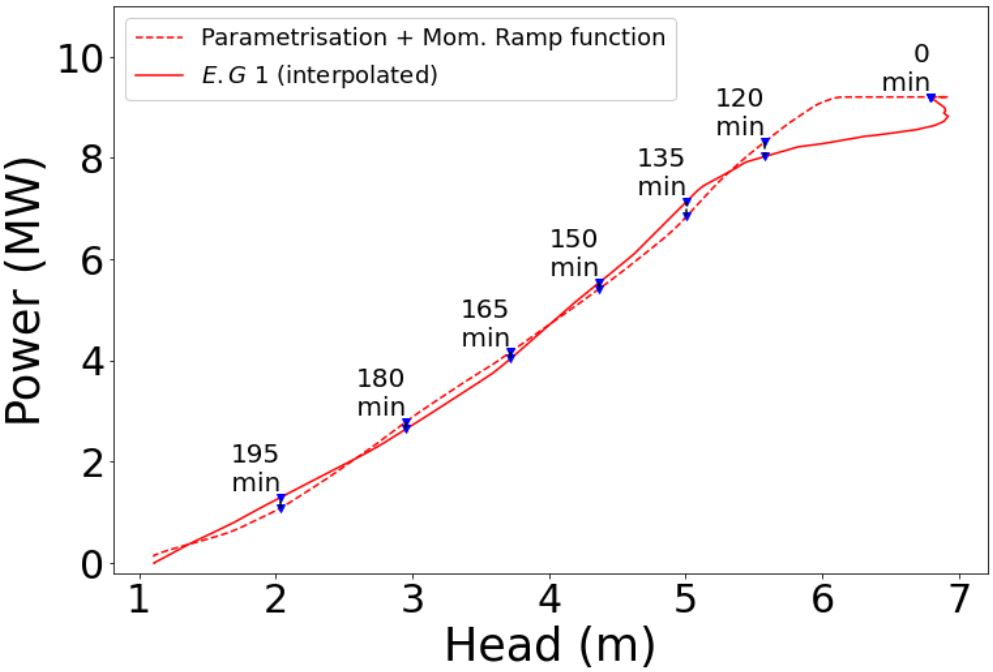}
    \caption{$E.G$ 1.} \label{EG1}
  \end{subfigure} \hfill \bigskip
  \begin{subfigure}[t]{.49\linewidth}
    \centering\includegraphics[width=\linewidth]{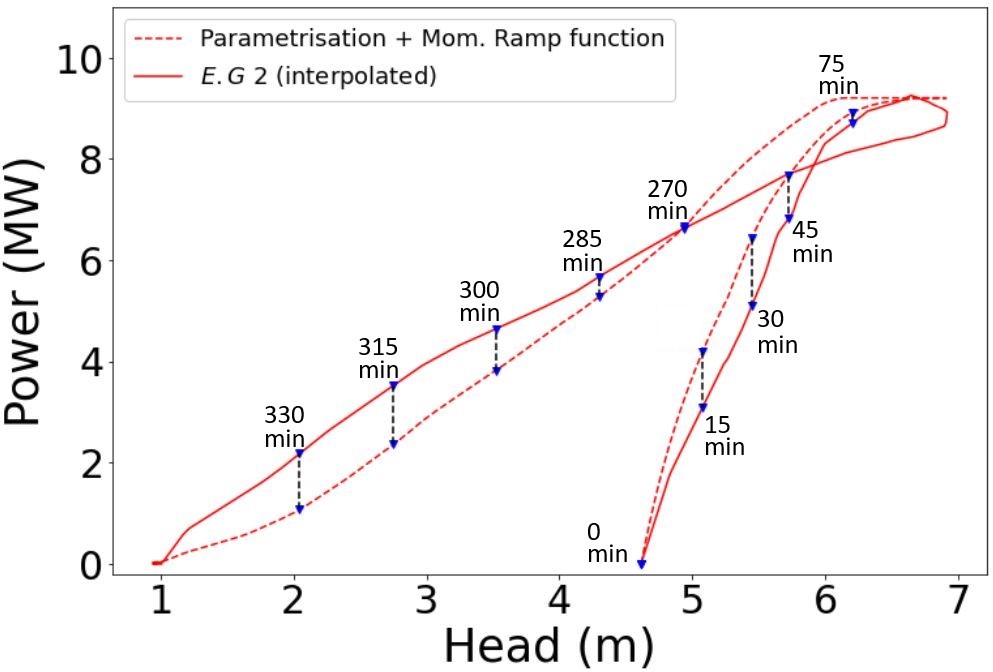}
    \caption{$E.G$ 2.}
    \label{EG2}
  \end{subfigure} \hfill \bigskip
  \begin{subfigure}[t]{.49\linewidth}
    \centering\includegraphics[width=\linewidth]{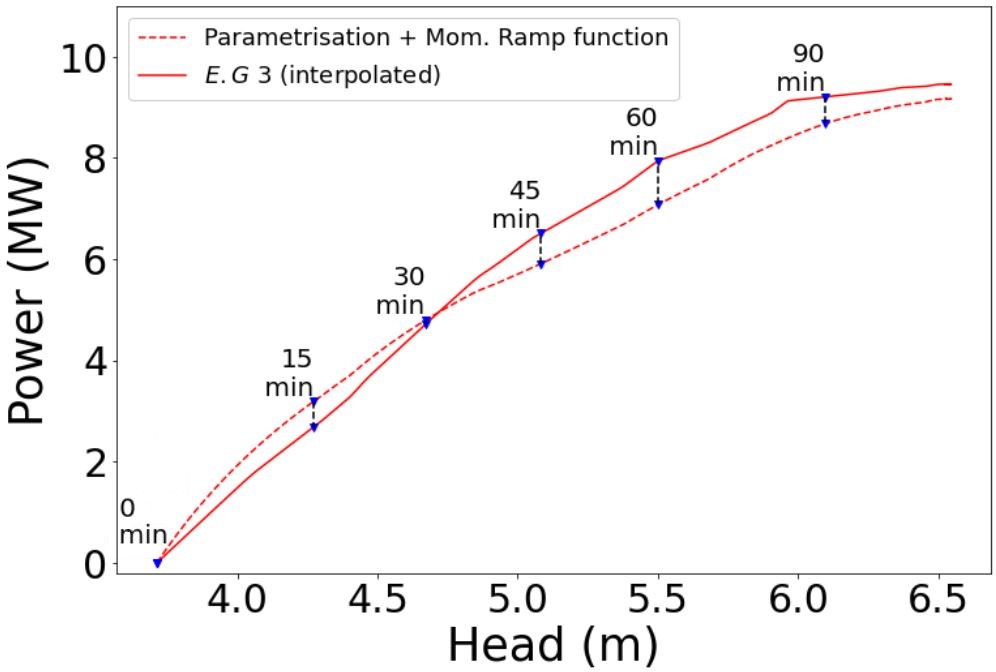}
    \caption{$E.G$ 3.} \label{EG3}
  \end{subfigure} \hfill
  \begin{subfigure}[t]{.49\linewidth}
    \centering\includegraphics[width=\linewidth]{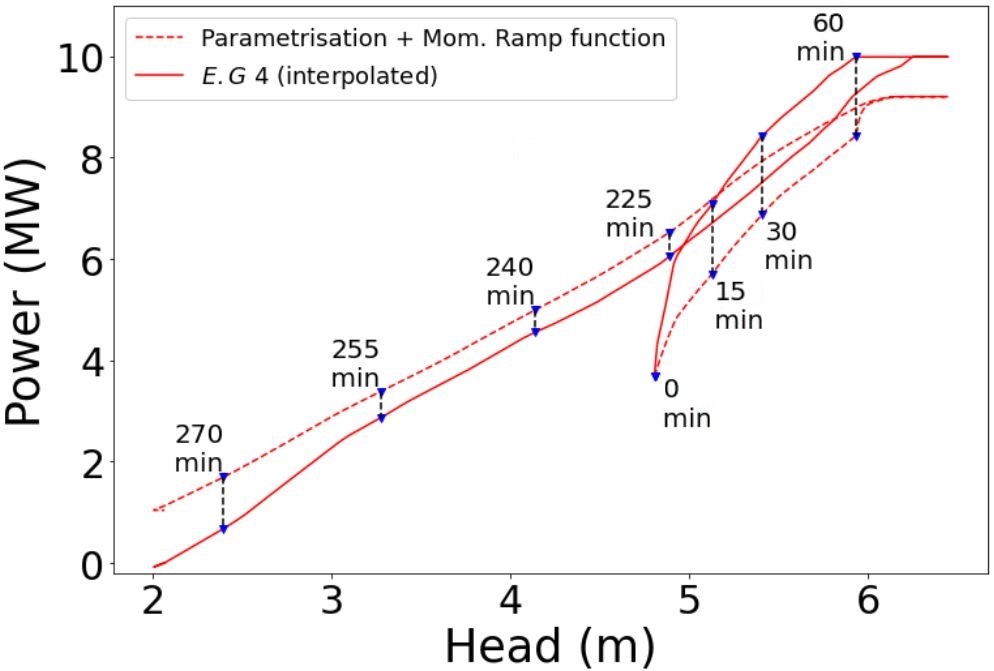}
    \caption{$E.G$ 4.}
    \label{EG4}
  \end{subfigure} \hfill \bigskip
  \begin{subfigure}[t]{.49\linewidth}
    \centering\includegraphics[width=\linewidth]{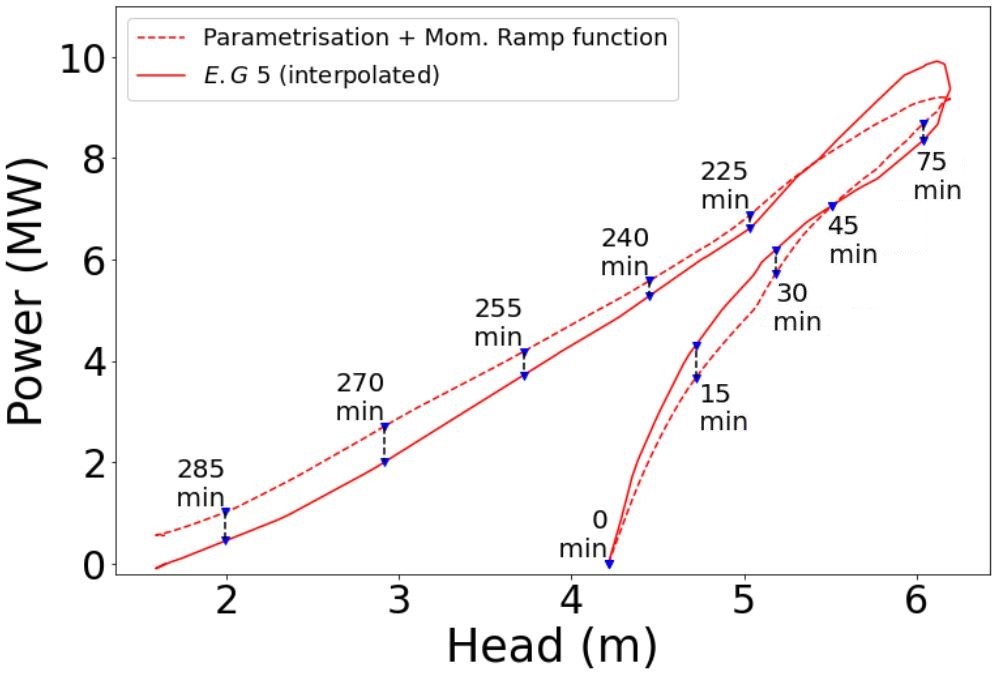}
    \caption{$E.G$ 5.}
    \label{EG5}
  \end{subfigure} \hfill
  \begin{subfigure}[t]{.49\linewidth}
    \centering\includegraphics[width=\linewidth]{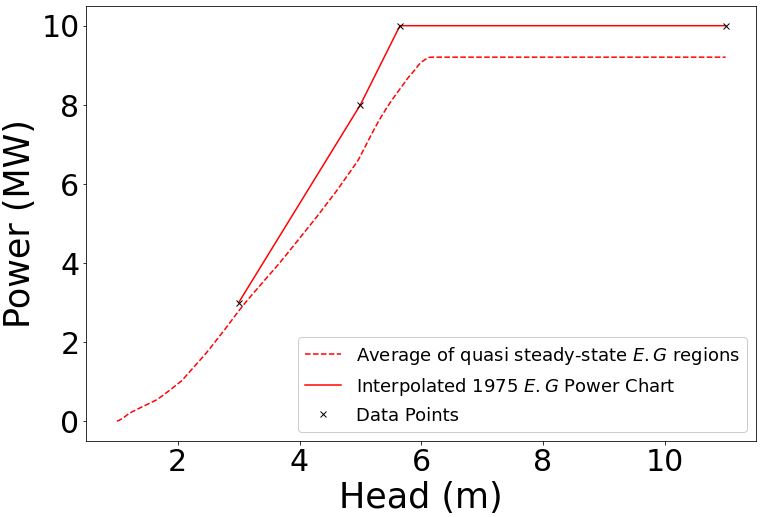}
    \caption{Parametrised (average of quasi steady-state $E.G$ regions) vs data interpolated hill charts.}
    \label{EGPC}
  \end{subfigure} \hfill  \bigskip
  \caption{($a - e$) Comparison between interpolated experimental data for power generation (during $E.G$) against our parametrised $E.G$ turbine hill chart, shown in ($f$), augmented with the momentum ramp function, for La Rance.} \label{EGParam}
\end{figure}

\begin{figure*}
  \centering
  \begin{subfigure}[t]{.49\linewidth}
    \centering\includegraphics[width=\linewidth]{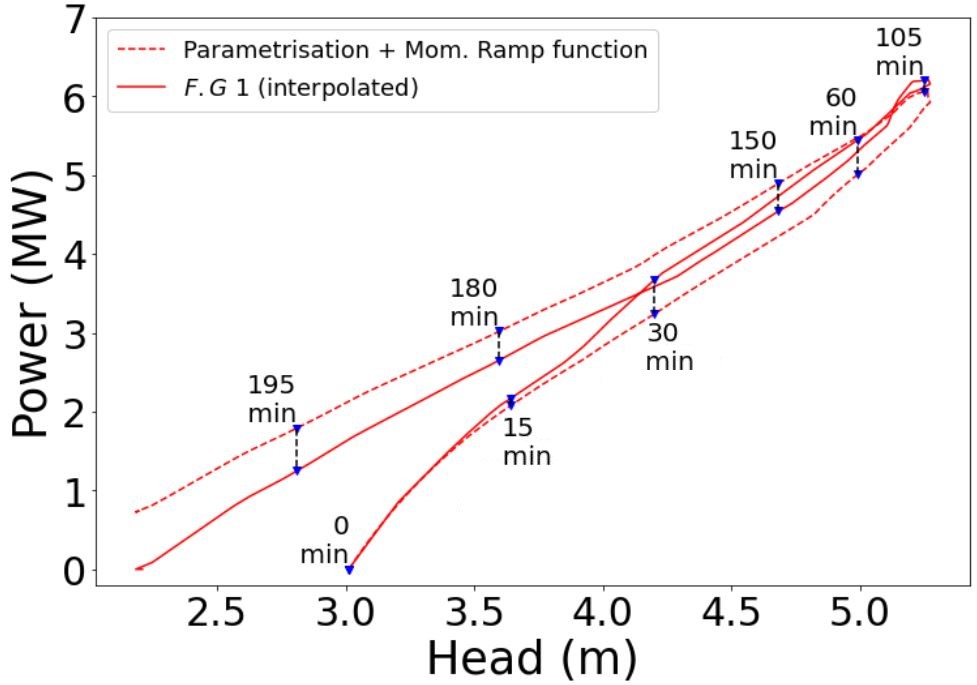}
    \caption{$F.G$ 1.} \label{FG1}
  \end{subfigure} \bigskip \hfill
  \begin{subfigure}[t]{.49\linewidth}
    \centering\includegraphics[width=\linewidth]{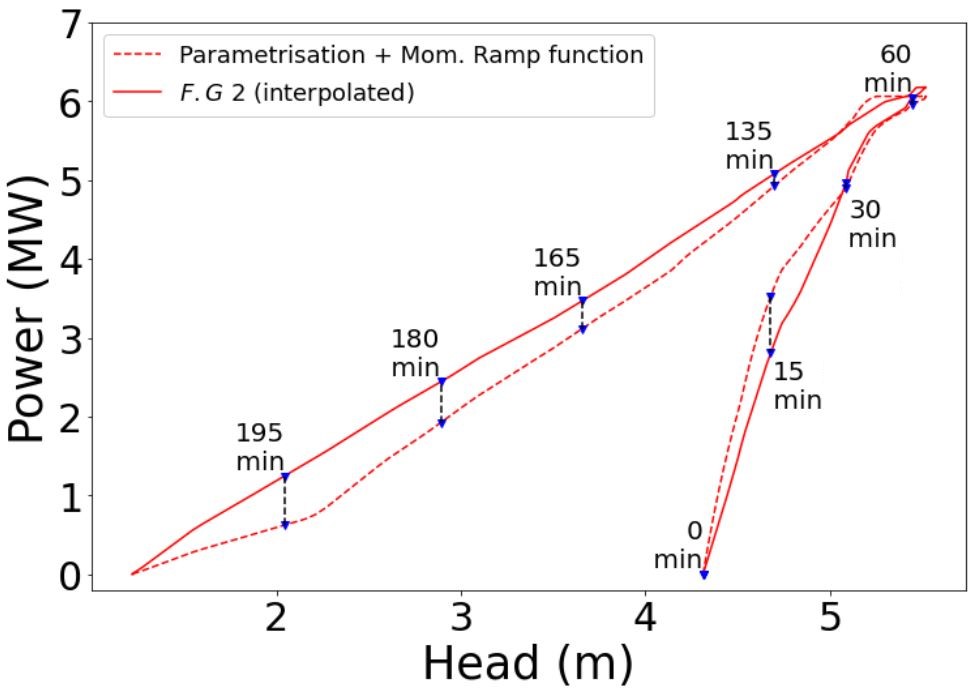}
    \caption{$F.G$ 2.}
    \label{FG2}
  \end{subfigure}
  \begin{subfigure}[t]{.55\linewidth}
    \centering\includegraphics[width=\linewidth]{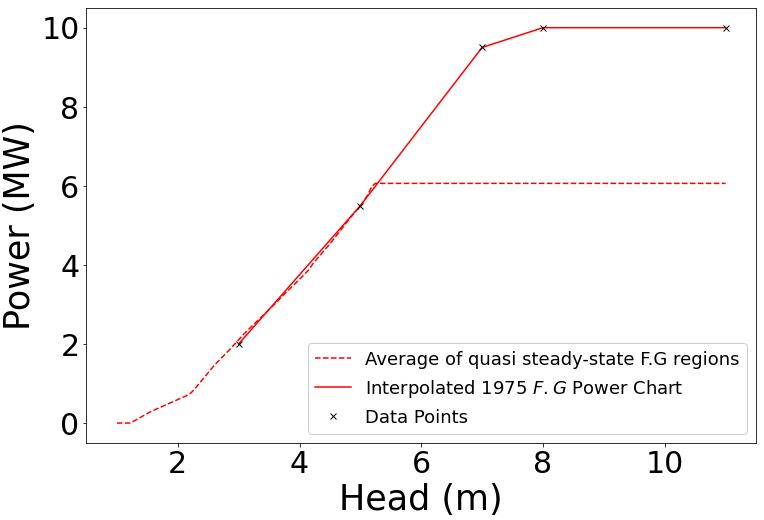}
    \caption{Parametrised (average of quasi steady-state $F.G$ regions) vs data interpolated hill charts.}
    \label{FGPC}
  \end{subfigure} \bigskip
  \caption{($a, b$) Comparison between interpolated experimental data for power generation (during $F.G$) against our parametrised $F.G$ turbine hill chart, shown in ($c$), augmented with the momentum ramp function, for La Rance.} \label{FGParam}
\end{figure*}

Given power production and water level variations for the ocean and impounded lagoon (for the numbered $E.G$ and $F.G$ turbine modes of operation in Fig \ref{TWPRance}), we can draw interpolated (parametric) curves for the turbine power output as a function of head, with time as hidden parameter: 
\begin{equation}
    P_{int} (EG_i) = P_{int}(h,t) \quad \text{or} \quad P_{int} (FG_i) = P_{int}(h,t),
\end{equation}
where ``$i$'' corresponds to the $i$th occurrence of $E.G$ and $F.G$ turbine modes. The interpolated $P_{int} (EG_i)$ and $P_{int} (FG_i)$ curves, for every ``$i$'' are shown as solid curves in Figs. \ref{EG1}, \ref{EG2}, \ref{EG3}, \ref{EG4}, \ref{EG5} and Figs. \ref{FG1}, \ref{FG2}, respectively. These figures showcase power production as a function of head difference, where time evolution is presented by labels over the interpolation. Considering scenarios where the starting phase for the turbines is available ($P_{int} (EG_2)$, $P_{int} (EG_3)$, $P_{int} (EG_5)$ and $P_{int} (FG_1)$, $P_{int} (FG_2)$), the interpolated results show a two-step process where initially (i) power outputs rapidly increases from a starting non-zero head difference until reaching a plateau (maximum head difference and power output). (ii) Then, from this plateau, water head variations, therefore power generation, slowly decrease with time, until power production is ceased. We interpret the first process as non-steady acceleration stage where the turbine, initially at rest, is submitted to a starting operational water head $H_{start}$, accelerating until maximum power generation is achieved. Conversely, the second process is interpreted as a quasi steady-state deceleration stage, where the turbine is submitted to a slowly decreasing water head, taking $\gtrsim 3$ hours to reach $H_{min}$. With this interpretation, we take the average of all available quasi steady-state phases for $E.G$ and $F.G$ modes, considering power output to be a pure function of head difference ``$h$'', thus obtaining hill charts ($P_{EG} (h)$, $P_{FG} (h)$) for both modes of operation. A comparison of our parametrised hill charts against interpolated hill charts from \cite{lebarbier1975power} is shown in Figs. \ref{EGPC} and \ref{FGPC}, for $E.G$ and $F.G$ modes of operation, respectively. The differences observed between these hill charts could be due to a series of factors not available to us, such as the pitching of runner blades/guide vanes. Nevertheless, proceeding with our parametrised hill charts enabled a satisfactory 0D simulation of turbines, as will be shown in the following sections.  

\begin{figure}[h]
	\centering
	\includegraphics[width=.7\linewidth]{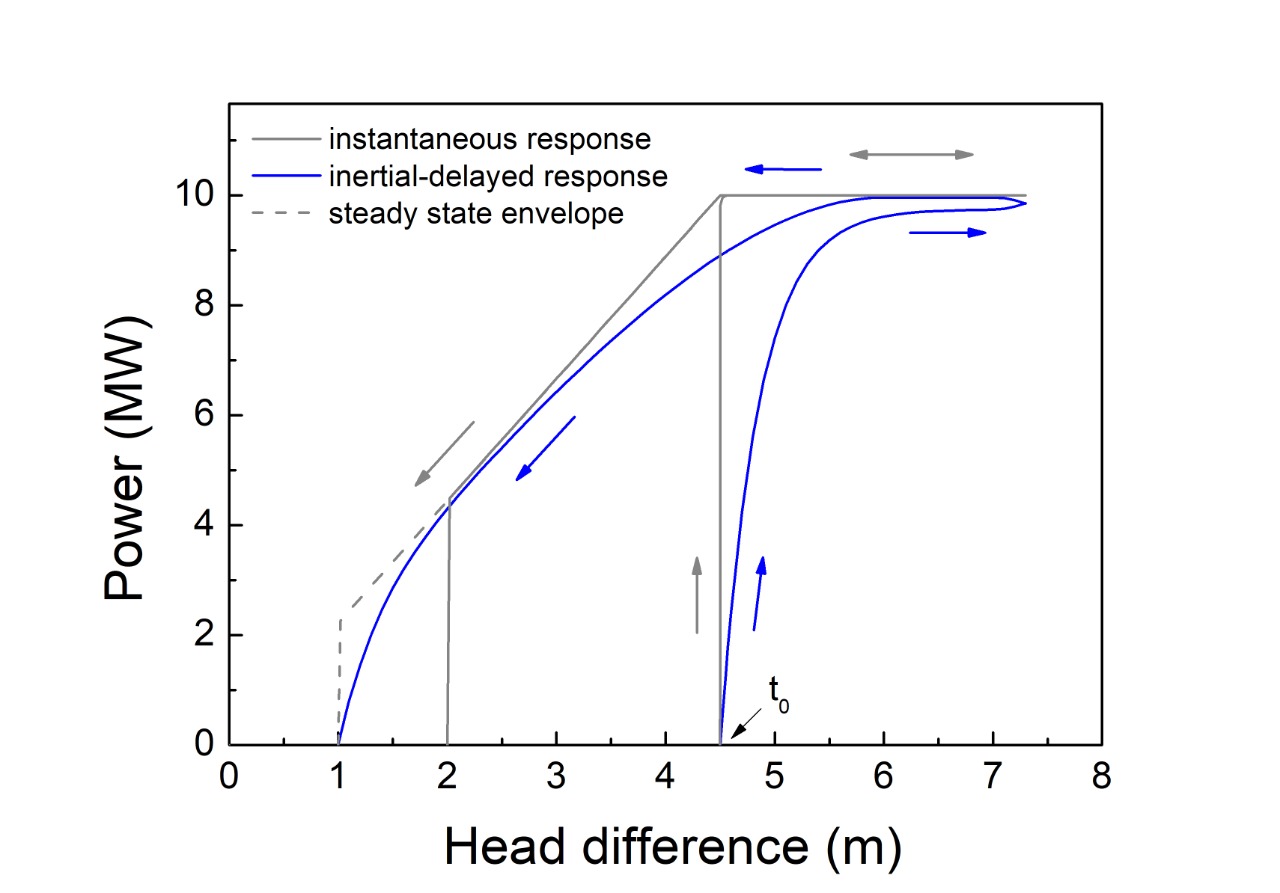}
	\caption{Sketch of augmenting the parametrised turbine hill chart (grey, instantaneous response) with the momentum ramp function (blue line). Time evolution, for starting turbine operation with a head difference $\approx 4.5~m$, is indicated with arrows.}
	\label{MomRampAug} 
\end{figure}

In order to simulate the smooth power output evolution for both accelerating and decelerating stages of turbine operation, the momentum ramp function, presented in Section \ref{MomRampS}, is utilised. A sketch showcasing the result of augmenting the parametrised hill chart with the momentum ramp function can be seen in Fig. \ref{MomRampAug}. 

For best adjusting the predicted power output (from Eq.~(\ref{MomRampPow})) as a function of water head, best $\zeta$ values for accelerating and decelerating stages of $E.G$ and $F.G$ modes of operation ($\zeta_{AE}$, $\zeta_{AF}$ and $\zeta_{DE}$, $\zeta_{DF}$ respectively) are found by minimising the Sum of Squares Residuals (Eq.~\ref{SSR}) between each interpolated hill chart ($P_{int} (EG_i)$, $P_{int} (FG_i)$) and the parametrised hill chart ($P_{EG} (h)$, $P_{FG} (h)$) augmented with the momentum ramp function (Eq.~\ref{MomRampPowEG}, \ref{MomRampPowFG}). It is worth noting that, when minimising the residuals, the water head input for both charts is also a function of time.
\begin{equation}\label{MomRampPowEG}
   \hat{P} (h, t+1, \zeta_{jE}) = P_{EG}(h) + \left(\hat{P} (h, t, \zeta_{jE}) - P_{EG}(h) \right)e^{-\Delta t/\zeta_{jE}},
\end{equation}
for $E.G$, and
\begin{equation}\label{MomRampPowFG}
   \hat{P} (h, t+1, \zeta_{jF}) = P_{FG}(h) + \left(\hat{P} (h, t, \zeta_{jF}) - P_{FG}(h) \right)e^{-\Delta t/\zeta_{jF}},
\end{equation}
for $F.G$, where ``$j = A$'' if $\hat{P}(t+1) > \hat{P}(t)$, else $j = D$.
\begin{equation}\label{SSR}
    SSR = \sum_{EG, FG} \sum_{t} (P_{int}(h(t),t) - \hat{P} (h(t),t,\zeta))^2.
\end{equation} 

Optimum $\zeta_{AE} = 14.2~min$, $\zeta_{AF} = 11.257~min$, $\zeta_{DE} = 1.355~min$ and $\zeta_{DF} = 1.091~min$ are then obtained, by only considering scenarios where the starting phase for the turbines were available, i.e. $EG_2$, $EG_3$, $EG_5$, $FG_1$ and $FG_2$.

Results of the parametrised hill chart augmented with the momentum ramp function with optimum $\zeta$ values are shown as dashed curves in Figs. \ref{EG1}, \ref{EG2}, \ref{EG3}, \ref{EG4}, \ref{EG5} and Figs. \ref{FG1}, \ref{FG2}. It is worth stressing that these results have been obtained with fixed sets [$\zeta_{AE}$, $\zeta_{DE}$] and [$\zeta_{AF}$, $\zeta_{DF}$] for $E.G$ and $F.G$ modes of operation, respectively.

\subsection{Equivalent Lagoon Wetted Area}\label{ALaRance}

So far in the literature, and to the best of our knowledge, 0D numerical simulations of TRS have utilised bathymetric data for estimating a ``flat'' lagoon wetted area for TRS. Although this technique can enable accurate simulation of TRS for ``small-scale'' systems, such as the SBL, as the impounded area and length increases (like in Severn barrage), this technique starts presenting significant deviations when compared to more realistic 2D models \cite{angeloudis2017comparison}.

In this section, we showcase a simple methodology that can be applied to any coastal region to extract an equivalent lagoon wetted area representation. This methodology was applied in our La Rance case study, since bathymetric data for the barrage was not available. In fact, knowing the real impounded area of La Rance is even more complicated, because, like any estuary, La Rance presents a severe 
accumulation of silt and sand due to siltation (a natural phenomenon due to sedimentation), amounting to approximately $50,000 ~m^3$ of sediment deposited each year in the estuary \cite{sediment,sediment2} -- altering La Rance´s bathymetry. In Section \ref{Val1} we show that the method proposed here enabled an accurate prediction of lagoon water level evolution when operating turbines (in power generation and pump modes) and sluices, even though the geographical maximum lagoon wetted area spans $22 ~km^2$ (twice that of SBL) and, given its narrow $720 ~m$ barrage, has an approximate length $>30 ~km$ \cite{rourke2010tidal}.

As shown in \cite{lisboa2017optimal}, analytical estimates for the available energy in a TRS can utilise the impounded volume of water instead of bathymetric representations. In a similar fashion, we utilise the stored water volume data shown in Table \ref{TidalPrism}, to derive an equivalent lagoon wetted area for the La Rance 0D model. In the literature, this equivalent area is attained from the inter-tidal prism volume estimate, which have been used to estimate the impounded water volume of rivers that end in tidal regions \cite{lakhan2003advances}. In its simplest form, a constant wetted area $Al$ is used, so the stored volume of water $\Delta V$ is proportional to variations in tidal level (tidal range) \cite{lakhan2003advances}. We note that this simplified model yielded fairly good results when applied to our La Rance 0D model. However, considering that there are several model approximations that can be used to estimate $Al$ as a function of tidal level variations \cite{d2010tidal}, we proceed our study with a linear approximation for $Al$ (which gave improved results). For the sake of simplicity, we use the tidal datum (height $z$ from the lowest equinoctial low tide, defined as $z = 0$), instead of the tidal prism. So that, 
\begin{equation} \label{linearA1}
    Al(z) = 2s \times z + Al_0,
\end{equation}
where $Al_0$ is the equivalent wetted area at $z = 0$ and $2s$ the slope. From this approximation, the stored volume ($\Delta V = V-V_0$) at a tidal datum $z$, obtained by integrating Eq.~ \ref{linearA1} from $0$ to $z$, becomes a second order polynomial:
\begin{equation}
    \Delta V = s \times z^2 + A_0 \times z.
\end{equation}

\begin{figure}[h]
  \centering
  \begin{minipage}[b]{.6\textwidth}
    \includegraphics[width=1\linewidth]{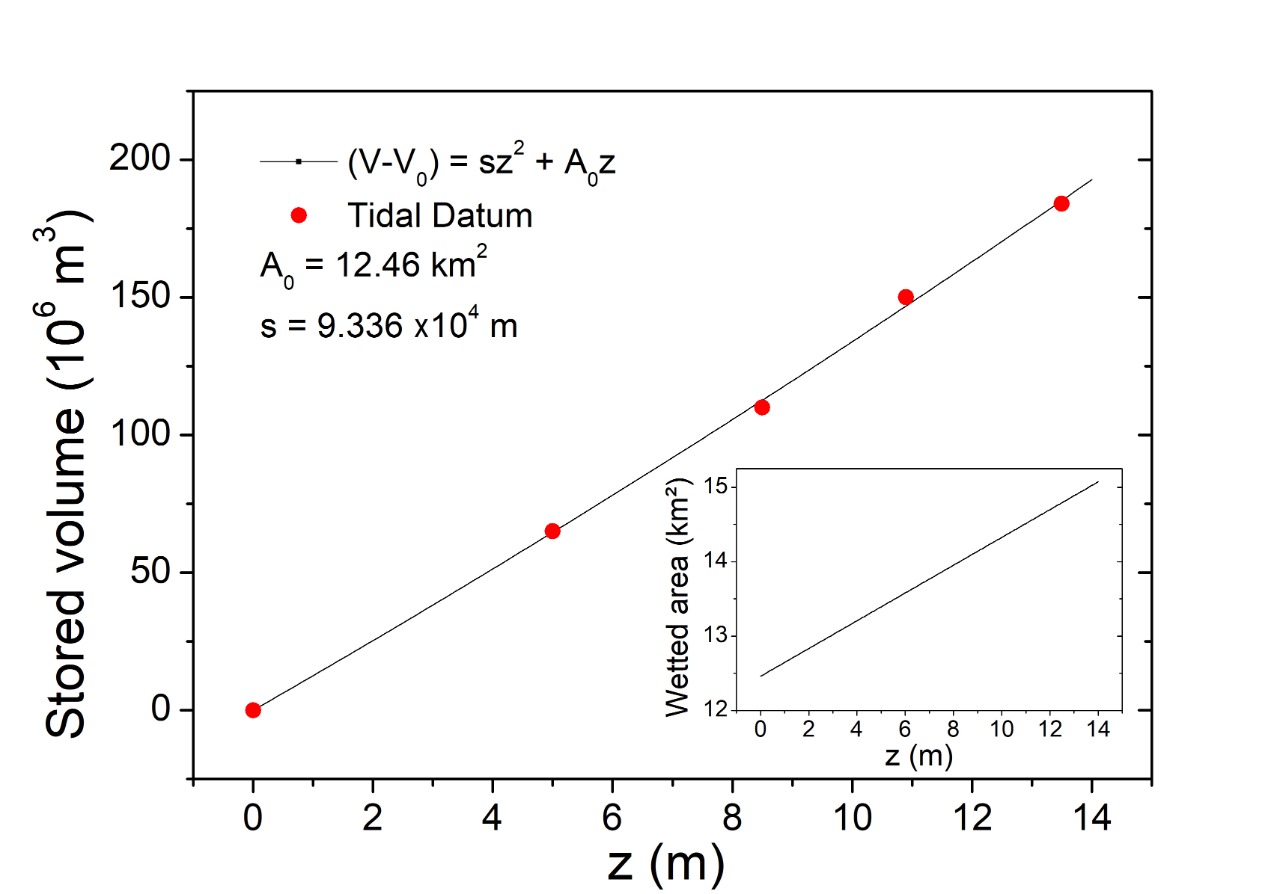}
  \end{minipage} \bigskip  \caption{Quadratic adjust of the volume of stored water ($\Delta V$), and linear equivalent wetted area estimate (insert) as a function of tidal datum $z$ for the La Rance estuary.} \label{LaRanceTP}
\end{figure}

Now the coefficients $s$ and $Al_0$ can be estimated with a least square method, utilising the point measurements ($z_i, \Delta V_i$) from Table \ref{TidalPrism}, as a target. The obtained quadratic fit for $\Delta V$ is shown in Fig. \ref{LaRanceTP}, together with found values for $s$ and $Al_0$. With these parameters, the wetted area given by Eq.~ \ref{linearA1} is also plotted and shown as an insert in Fig. \ref{LaRanceTP}.

\subsection{Sluicing}\label{SLaRance}

\begin{figure}[h]
  \centering
  \begin{subfigure}[t]{.47\linewidth}
    \centering\includegraphics[width=\linewidth]{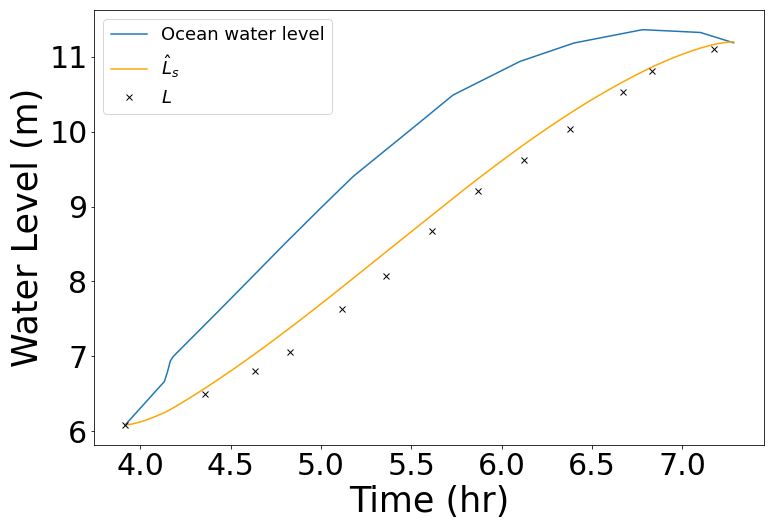}
    \caption{Sluicing Region 1.} \label{s1}
  \end{subfigure}  \hfill \bigskip
  \begin{subfigure}[t]{.48\linewidth}
    \centering\includegraphics[width=\linewidth]{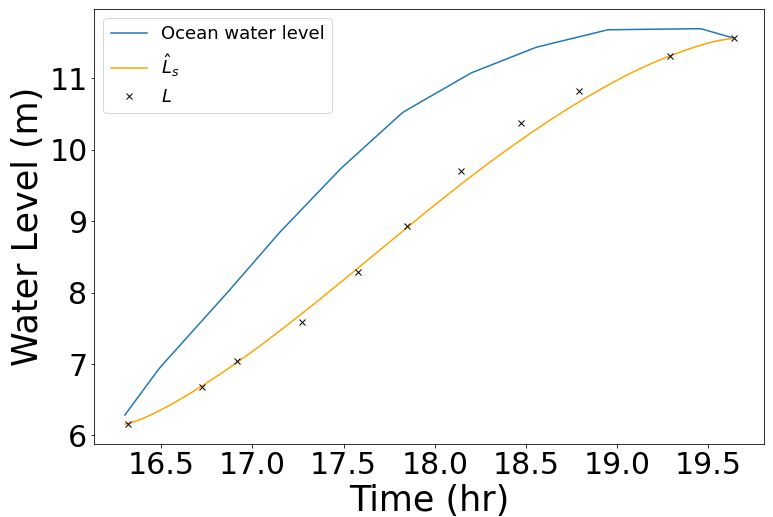}
    \caption{Sluicing Region 2.}
    \label{s2}
  \end{subfigure}
  \begin{subfigure}[t]{.49\linewidth}
    \centering\includegraphics[width=\linewidth]{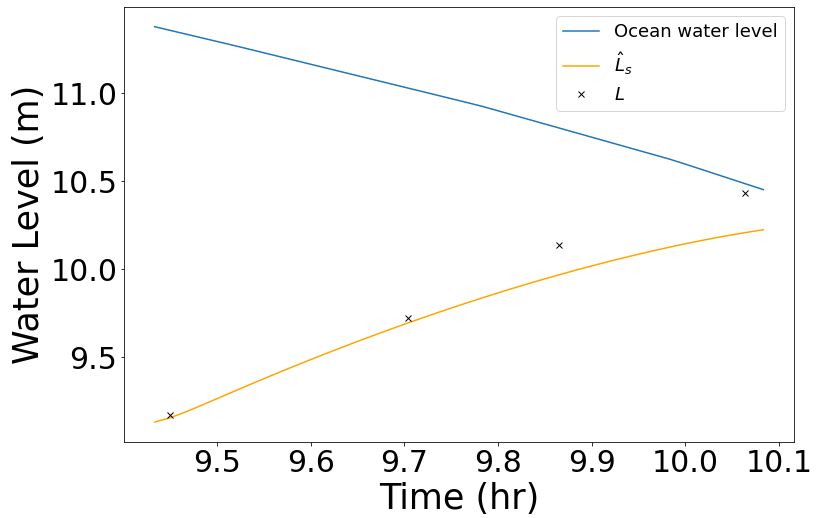}
    \caption{Sluicing Region 3.} \label{s3}
  \end{subfigure} \hfill \bigskip
  \begin{subfigure}[t]{.49\linewidth}
    \centering\includegraphics[width=\linewidth]{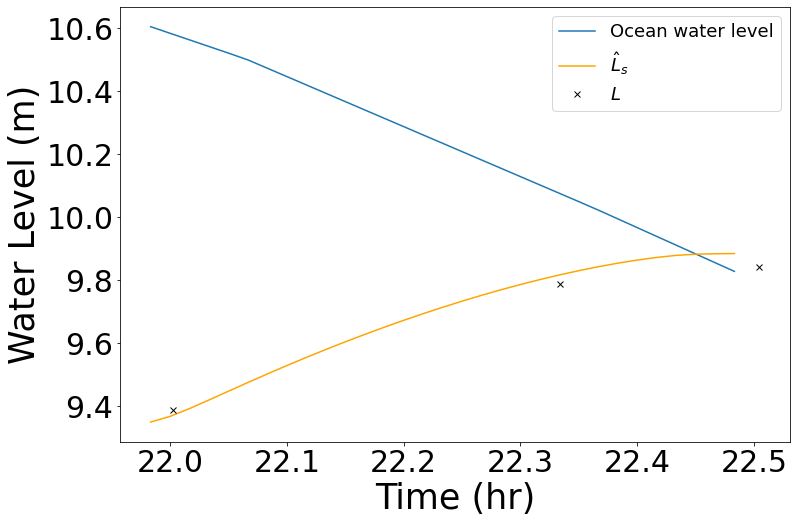}
    \caption{Sluicing Region 4.}
    \label{s4}
  \end{subfigure}
  \caption{Comparison between predicted ($\hat{L}_s$) and measured ($L$) lagoon water level variations during sluicing stage for La Rance. Measured experimental data from \cite{lebarbier1975power} is shown in crosses.} \label{sluicingSt}
\end{figure}

For parametrising the sluicing stages shown in Fig. \ref{EGRwl} and \ref{EGRPow}, we need to define appropriate (i) $\zeta$ values for our momentum ramp function (Eq.~\ref{MomRamp}) and (ii) discharge coefficient ($C_d$) for the orifice equation (Eq.~\ref{Orif}), responsible for estimating flow rates from the hydraulic structures.

In contrast with turbines (in power generation mode) that have accelerating and decelerating stages (Section \ref{TLaRance}), sluices allow for the free passage of water with minimum resistance, independent if head differences are increasing or decreasing. Due to this symmetric behaviour, we assume a constant $\zeta_s$ value for our momentum ramp function estimate. Since we do not have measured data to fit a best $\zeta_s$ value for sluices, we adopt the lower bound $\zeta_s = 1.091~min$. 

Beyond sluice gates, the sluicing stages shown in Fig. \ref{EGRwl} and \ref{EGRPow}, need also to account for turbine flow rates operating in ``idling'' mode. For simplification purposes, we assume that turbine operation in idling mode have a similar behaviour to sluices, since there is no electrical energy conversion. Therefore, the same $\zeta_s = 1.091~min$ is used. This similar behaviour is also supported by the literature, since flow rate estimates for sluices and idling turbines utilise the same ``orifice'' equation \cite{angeloudis2016numerical, aggidis2013operational, falconer2009severn}.

For best fitting Eq.~\ref{Orif} to La Rance, appropriate $C_d$ and $C_{dt}$, for sluices and idling turbines, respectively, need to be found. For sluice gates, we bound the $C_d$ estimate between two experimental values from the literature. As a lower bound, we have $C_d = 1$, from experimental results of a large sluice gate experiment by \cite{sellin1981severn}. As an upper bound, we utilise measurements from \cite{lebarbier1975power}, which indicate that, at its maximum, sluice gates from La Rance provide a flow rate of $9600 ~m^3/s$. Since the total sluice gate area for La Rance is $900 ~m^2$, we attain an upper bound of $C_d = 1.077$.

Similarly, idling turbines flow rate is bound by experimental measurements, at La Rance, from \cite{rolandez2014discharge}. In this work, flow rate measurements from a turbine unit under a fixed $4~m$ head was recorded in the range $182.2 ~m^3/s$ to $280 ~m^3/s$. From this data, upper and lower bound discharge coefficients for turbines and sluices are defined as:
\begin{equation}
\label{CdCdtbound}
[1 \leq C_d \leq 1.077] \text{ and } [.91 \leq C_{dt} \leq 1.4].
\end{equation}

For choosing best $C_d$ and $C_{dt}$ values within bounds (Eq.~\ref{CdCdtbound}), we utilise the 0D model for La Rance (Eq.~\ref{Num0D}), with the equivalent lagoon wetted area derived in the previous section, to predict lagoon water level variations during each sluicing stage ``$S$''. The predicted levels $\hat{L}_s$ are compared with measured data $L$ and a time-normalised sum of squared residuals (NSSD) is calculated:
\begin{equation}
    NSSD = \sum_{S = 1}^{S = 4} \frac{(\hat{L}_s - L)^2}{t_S}
\end{equation}

The best $C_d$ and $C_{dt}$ that minimise $NSSD$ for all sluicing stages are $C_d = 1.017$ and $C_{dt} = .967$. However, since adopting $C_d = C_{dt} = 1$ does not change results significantly, for the sake of simplification, we resume our work with a discharge coefficient of ``$1$'', for both sluices and idling turbines. Comparison between $\hat{L}_s$ and $L$ results, with $C_d = C_{dt} = 1$, are shown in Fig. \ref{sluicingSt}.

\subsection{Turbines -- Pumping Mode}\label{PLaRance}

As shown in Section \ref{StateTrs}, current idealised pump models for TRS (Eq.~\ref{PumpFlow}) fail to explain expected behaviours such as maximum pump shutoff head $h_s$ (when $Q_p = 0$), maximum pump flow rates $Q_M$ (for $h_p = 0$) and pump operation under positive head scenarios (aided by gravity) -- all observed in La Rance measurements \cite{lebarbier1975power}. For contemplating these behaviours, we utilise pump affinity laws in order to derive a general formulation for the pump flow rate as a function of head difference and applied power input: $Q_p = Q_p(h_p,P_{in})$. The affinity laws express the mathematical relationships between several variables involved in the performance of kinetic pumps (centrifugal or axial). These laws show that, under dynamically similar conditions, dimensionless parameters remain constant. They are useful for predicting pump performance changes when varying either (i) pump operational speed or (ii) pump impeller diameter \cite{stewart2018surface}. Since bulb turbines operating in reverse have a similar behaviour to axial-flow pumps \cite{arshenevskii1979characteristics}, we continue this section with the assumption that affinity laws can be applied to bulb turbines in pump mode.

Each affinity law postulate can be expressed as a set of three equations, where pump flow rate $Q_p$, negative head $h_p$ and input power $P_{in}$ are estimated for a point ``$p2$'' using a known point ``$p1$'' as reference. Then, following the affinity laws, values for ``$p2$'' are obtained as a function of either pump rotation ($N$) or pump impeller diameter ($D$), following the first or second postulate from the affinity laws, respectively. From the first postulate \cite{stewart2018surface}:
\begin{equation} 
  \frac{Q_{p1}}{Q_{p2}} = \frac{N_1}{N_2}; \quad \frac{h_{p1}}{h_{p2}} = \left (\frac{N_1}{N_2} \right )^2; \quad \frac{P_{in1}}{P_{in2}} = \left (\frac{N_1}{N_2} \right )^3.
  \label{AffLaw1}
\end{equation}

For the relationship shown in Eq.~\ref{AffLaw1} to be true, pump efficiency must remain relatively constant as we move from ``$p1$'' to ``$p2$'', for a fixed diameter pump \cite{stewart2018surface}. Utilising the first postulate of pump affinity laws (and assuming a near constant efficiency for any change in rotational speed) is sufficient for estimating a general pump equation for La Rance. The first postulate can also be manipulated, so that $Q_p$ and $h_p$ are obtained as a function of power input variation:
\begin{equation} 
  Q_{p1} = Q_{p2} \left( \frac{P_{in1}}{P_{in2}} \right)^{1/3},
  \label{QP}
\end{equation}
\begin{equation} 
  h_{p1} = h_{p2} \left( \frac{P_{in1}}{P_{in2}} \right)^{2/3}.
  \label{HP}
\end{equation}

For both $E.P$ and $F.P$ operational modes, the second order approximation for flow rate $Q_p(h_p,P_{in})$ as a function of head for a fixed (maximum) power input of $P_{in} = 6MW$, shown in Fig. \ref{EbbPumpFlow}, is utilised. For simplification purposes, $\{P_{in} = 6MW\} = P_6$. Also, since pump operational head $h_p$ depends on power input (Eq.~\ref{HP}), we have for the quadratic approximation $Q_p(h_p(P_6),P_{in} = P_6) = Q_p(h_p (P_6))$, so that:
\begin{equation} 
  Q_p(h_p(P_6)) = ah_p^2(P_6) + bh_p(P_6) + Q_M(P_6),\ \text{where} \ h_p(P_6)<=0.
  \label{QuadraticApprox}
\end{equation}
Eq.~\ref{QuadraticApprox} expresses a characteristic pump curve, where $Q_M(P_6)$ is the maximum pump flow rate expected, for $h_p = 0$ and $P_{in} = 6MW$. Knowing that $Q_p$ is zero at the shutoff head ($h_s$), the quadratic approximation can be expressed in terms of its roots, yielding a convenient form:
\begin{equation} 
  Q_p(h_p (P_6)) = a[h_p(P_6) - h_s(P_6)][h_p(P_6) - Q_M(P_6)/(a h_s(P_6))].
  \label{QuadraticApprox2}
\end{equation}

For deriving a characteristic pump curve for any power input ($P_{in}$) and pump head ($h_p(P_{in})$), we expand all terms in Eq.~\ref{QuadraticApprox2}, by utilising Eq.~\ref{HP} and Eq.~\ref{QP}, thus obtaining Eq.~\ref{Ratios}. For convenience, $(6MW/P_{in}) = R$.
\begin{equation}
\label{Ratios}
\begin{aligned}
Q_p(h_p (P_6)) &= Q_p(h_p (P_{in})) R^{1/3},  \\ 
Q_M(P_6) &= Q_M(P_{in}) R^{1/3},  \\
h_p (P_6) &= h_p (P_{in}) R^{2/3},  \\
h_s (P_6) &= h_s (P_{in}) R^{2/3}. 
\end{aligned}
\end{equation}

Performing the substitution of terms from Eq.~\ref{Ratios} into Eq.~\ref{QuadraticApprox2}, returns
\begin{equation} 
  Q_p(h_p, P_{in}) = aR[h_p(P_{in}) - h_s(P_{in})][h_p(P_{in}) - Q_M(P_{in})/(aR h_s(P_{in}))].
  \label{QuadraticApprox3}
\end{equation}

\begin{figure}
  \centering
  \begin{subfigure}[t]{.49\linewidth}
    \centering\includegraphics[width=\linewidth]{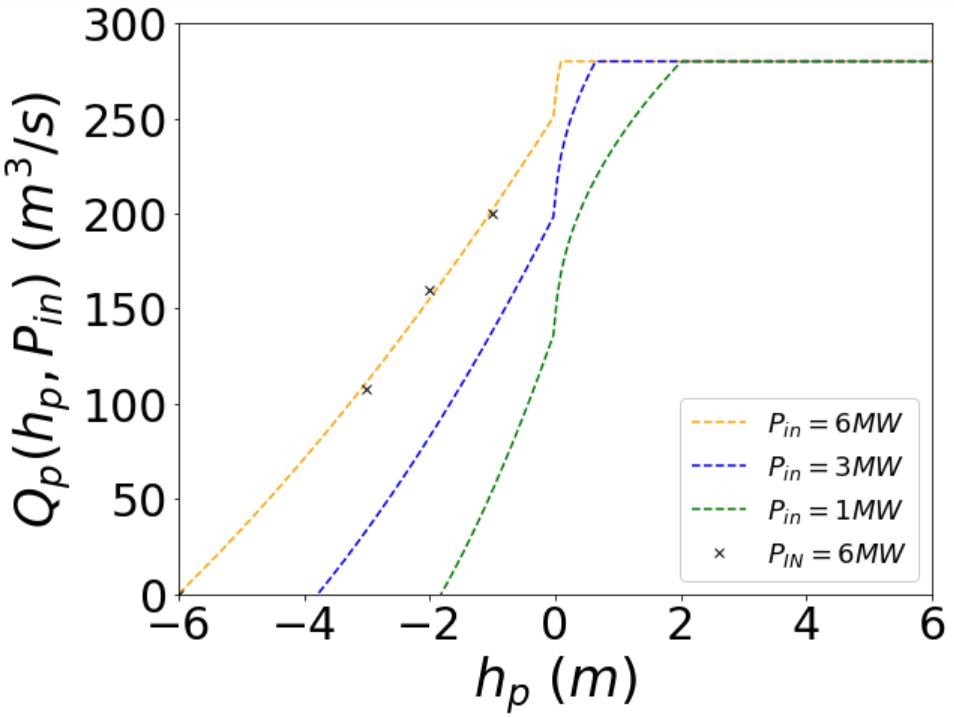}
    \caption{$E.P$.} \label{EPm}
  \end{subfigure} \hfill \bigskip
  \begin{subfigure}[t]{.49\linewidth}
    \centering\includegraphics[width=\linewidth]{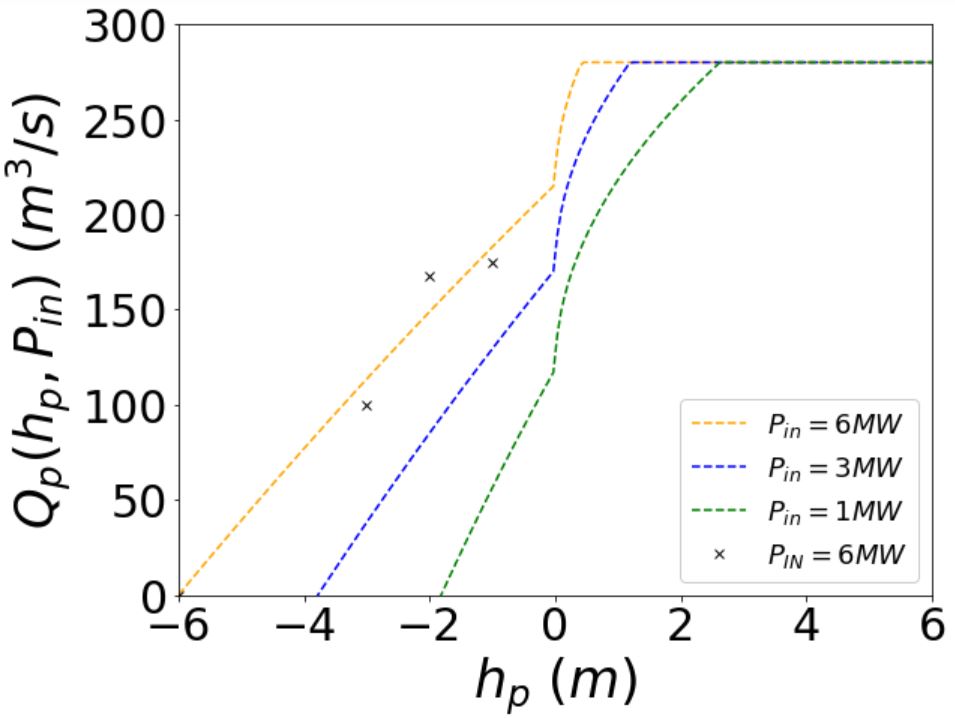}
    \caption{$F.P$.}
    \label{FPm}
  \end{subfigure} 
  \caption{La Rance $Q_p(h_p, P_{in})$ estimate for $E.P$ and $F.P$ turbine flow rates, for $P_{in} = 6, 3$ and $1$MW.} \label{EPandFP}
\end{figure}

\begin{figure}[h!]
  \centering
  \begin{subfigure}[t]{.47\linewidth}
    \centering\includegraphics[width=\linewidth]{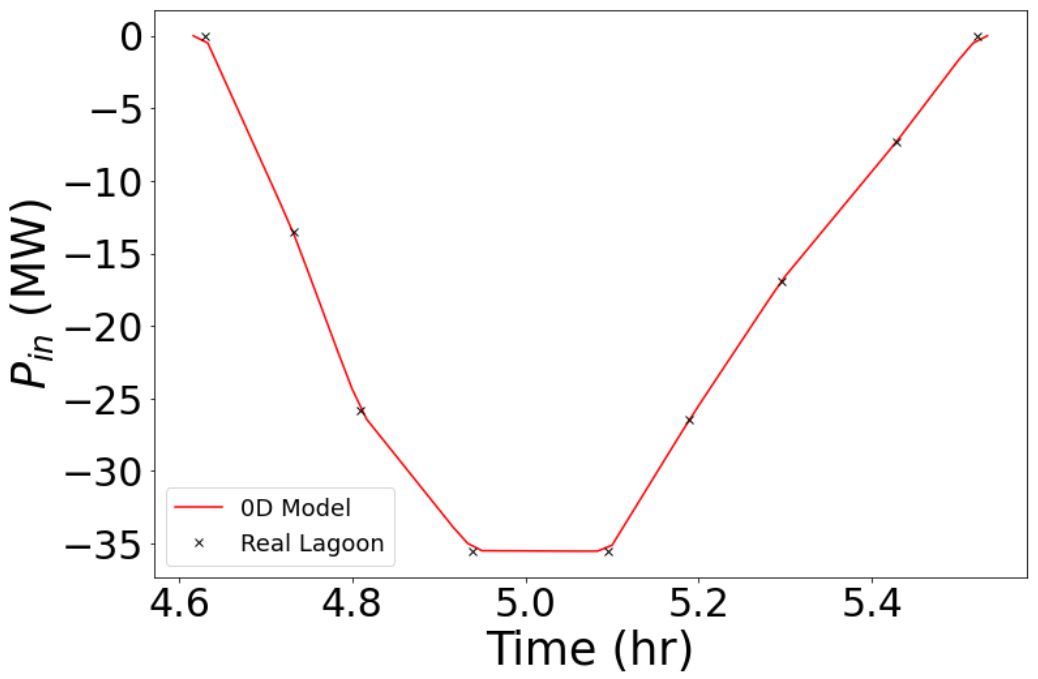}
    \caption{$P_{in}$ for $E.P$ 1.} \label{EP1Pow}
  \end{subfigure} \hfill \bigskip
  \begin{subfigure}[t]{.47\linewidth}
    \centering\includegraphics[width=\linewidth]{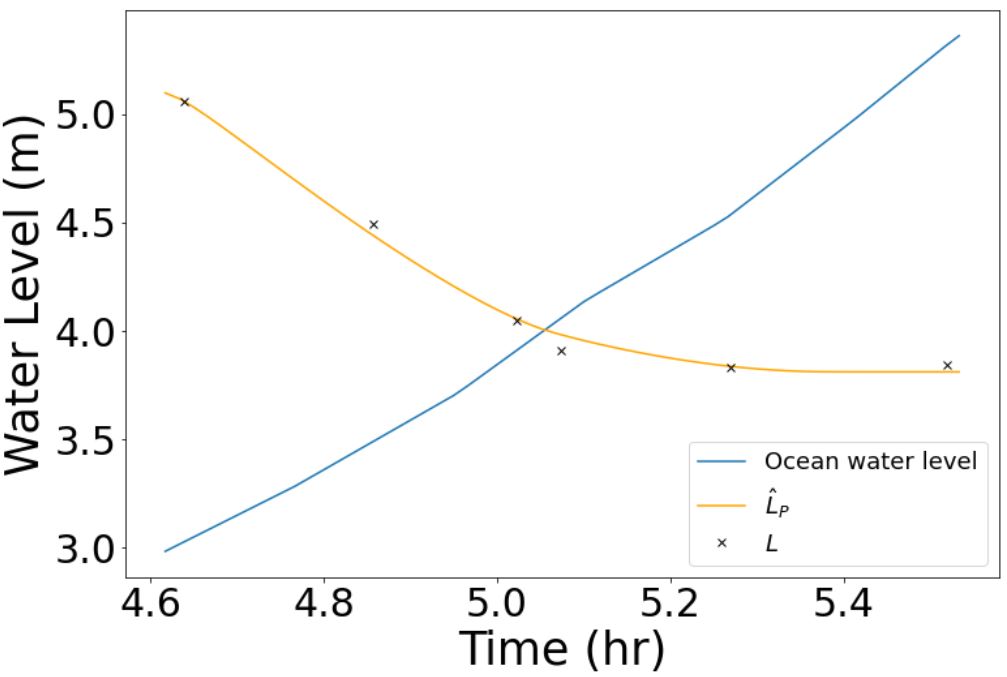}
    \caption{$\hat{L}_P$ and $L$ for $E.P$ 1.} \label{EP1Lwl}
  \end{subfigure}  \hfill
  \begin{subfigure}[t]{.47\linewidth}
    \centering\includegraphics[width=\linewidth]{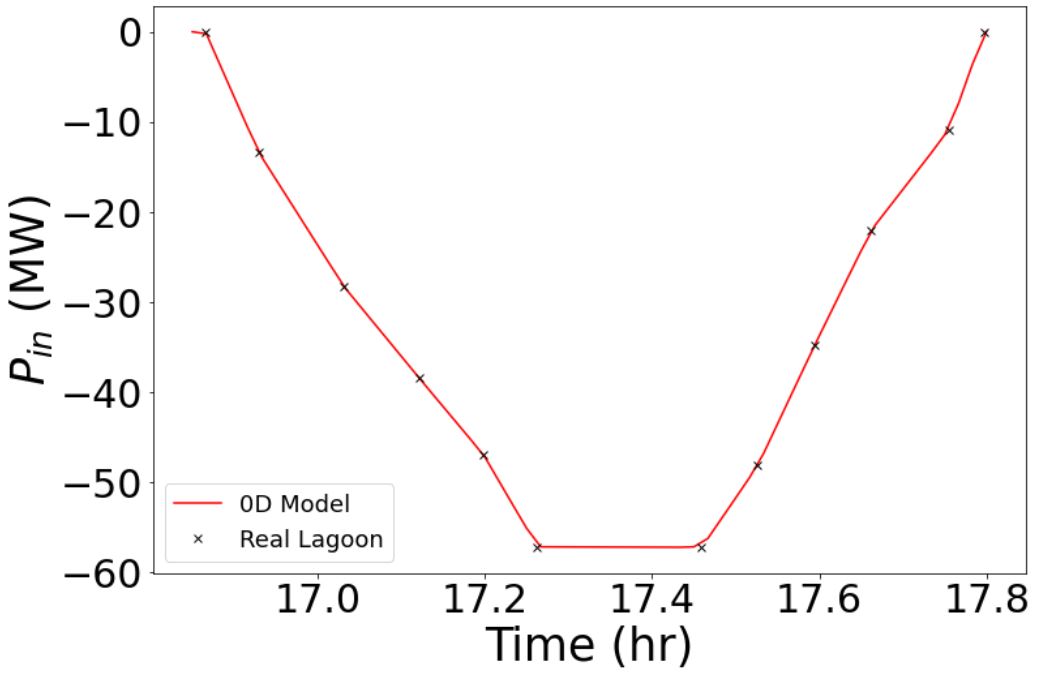}
    \caption{$P_{in}$ for $E.P$ 2.} \label{EP2Pow}
  \end{subfigure}  \hfill \bigskip
  \begin{subfigure}[t]{.47\linewidth}
    \centering\includegraphics[width=\linewidth]{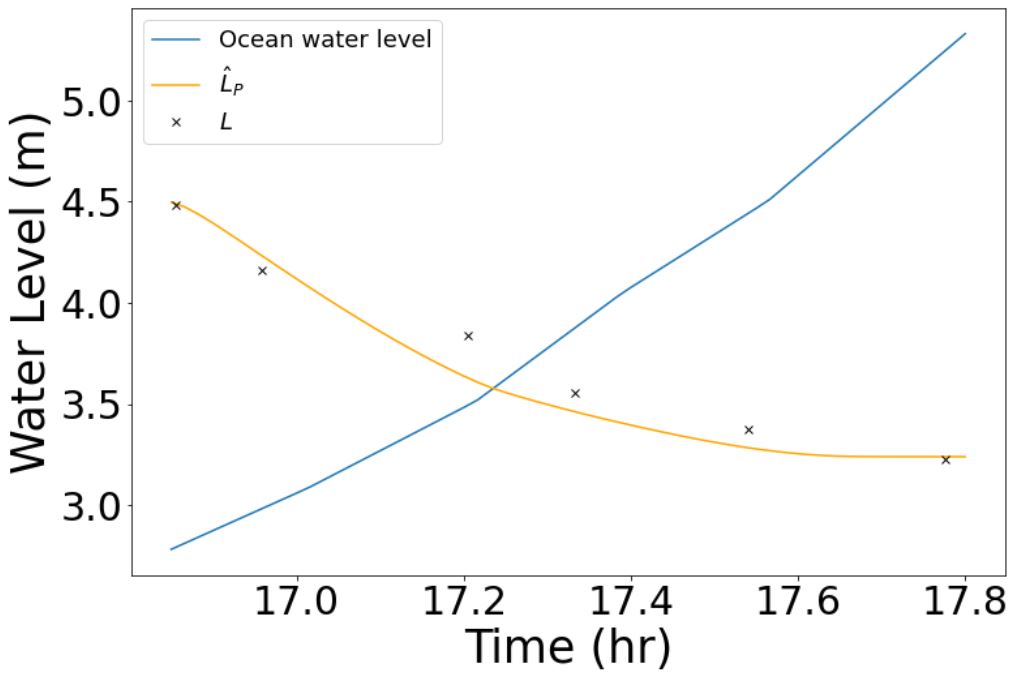}
    \caption{$\hat{L}_P$ and $L$ for $E.P$ 2.} \label{EP2Lwl}
  \end{subfigure}  \hfill
  \begin{subfigure}[t]{.47\linewidth}
    \centering\includegraphics[width=\linewidth]{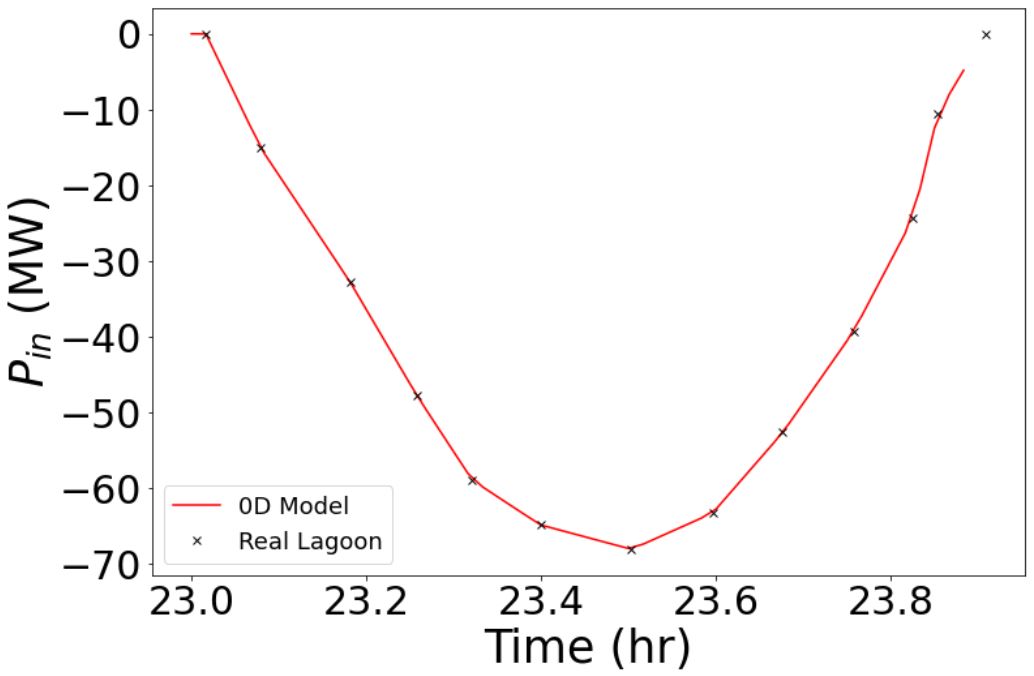}
    \caption{$P_{in}$ for $F.P$.} \label{FPPow}
  \end{subfigure} \hfill
  \begin{subfigure}[t]{.47\linewidth}
    \centering\includegraphics[width=\linewidth]{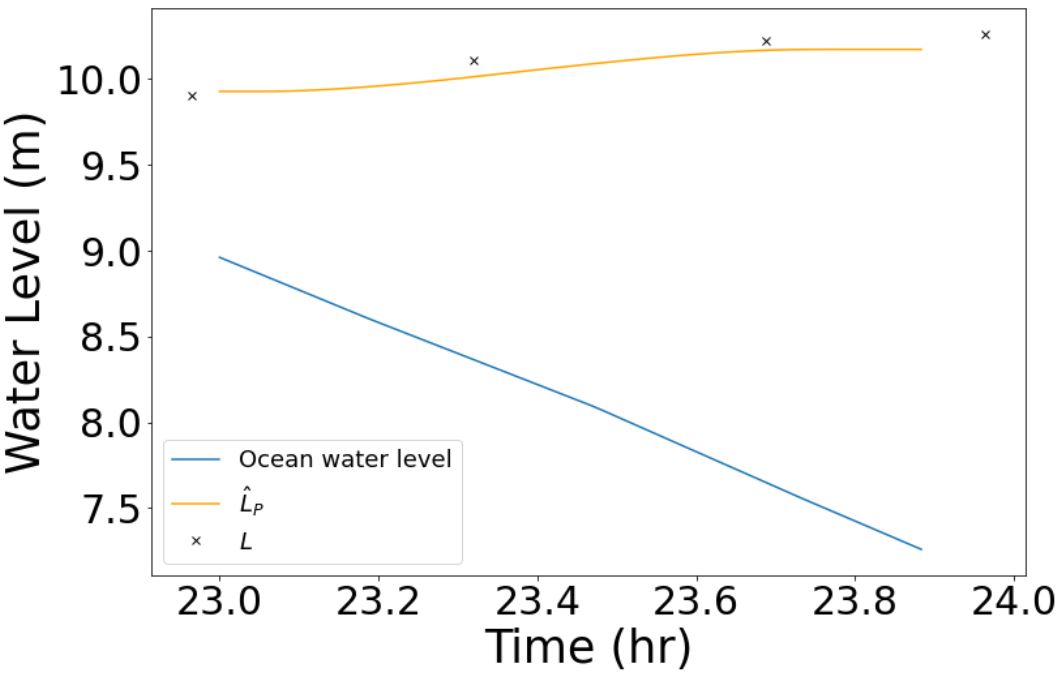}
    \caption{$\hat{L}_P$ and $L$ for $F.P$.} \label{FPLwl}
  \end{subfigure}
  \caption{Comparison between predicted $\hat{L}_P$ (yellow solid curves) and measured $L$ ($\times$) lagoon water level variations during pumping stages ($b, d, f$), along with measured ocean water levels (blue solid curves) and power input $P_{in}$ (red solid curves and black crosses in $a, c$ and $e$), used as inputs for the 0D La Rance model.} \label{pumpingSt}
\end{figure}

Eq.~\ref{QuadraticApprox3} allows for estimating pump flow rates for any given $h_p$ and $P_{in}$, including regions of pump shutoff head $h_s$ and maximum pump flow rate $Q_M$. Finally, in order to estimate pump flow rates for positive head scenarios, we assume that the maximum $Q_M$ attained for $h_p = 0$ is summed with the gravitational flow rate estimate (orifice equation for turbines) shown in Section \ref{SLaRance}. The maximum flow rate allowed during pumping (upper bound ``$ub$'') is $ub = 280 m^3/s$, as defined in Section \ref{ParamDef}. Examples of predicted pump flow rates, for $E.P$ and $F.P$ modes of operation are shown in Fig. \ref{EPandFP}.
For verifying the quality of our pump $Q_p(h_p, P_{in})$ estimate (Eq.~\ref{QuadraticApprox3}), we utilise the 0D model for La Rance (Eq.~\ref{0D}), with the equivalent lagoon wetted area derived in Section \ref{ALaRance}, to predict lagoon water level variations $\hat{L}_P$ during each $E.P$ and $F.P$ pumping stage. The predicted $\hat{L}_P$ values, for a given (measured) power input $P_{in}$ and varying ocean levels, are compared with measured lagoon water levels $L$ and shown in Fig. \ref{pumpingSt}.
As shown in Fig. \ref{pumpingSt}, the derived pump model for La Rance presented good agreement of results when predicting lagoon water level variations against measured data. These results were obtained considering a constant (lower bound) value for the momentum ramp function $\zeta_p = 1.091~min$.

The required steps to generalise the developed pump equation for bulb turbines of various diameters and power capacity are presented in the Supplementary Material.

\section{Validation of the Parametrised 0D La Rance Model} \label{Val1}

With the developed parametric models for turbines (in power generation and pump modes), sluices, equivalent lagoon area and momentum ramp function, we can now verify the accuracy of our 0D La Rance model in predicting lagoon water level variations and power output.

\begin{figure}[h!]
  \centering
  \begin{subfigure}[t]{.49\linewidth}
    \centering\includegraphics[width=\linewidth]{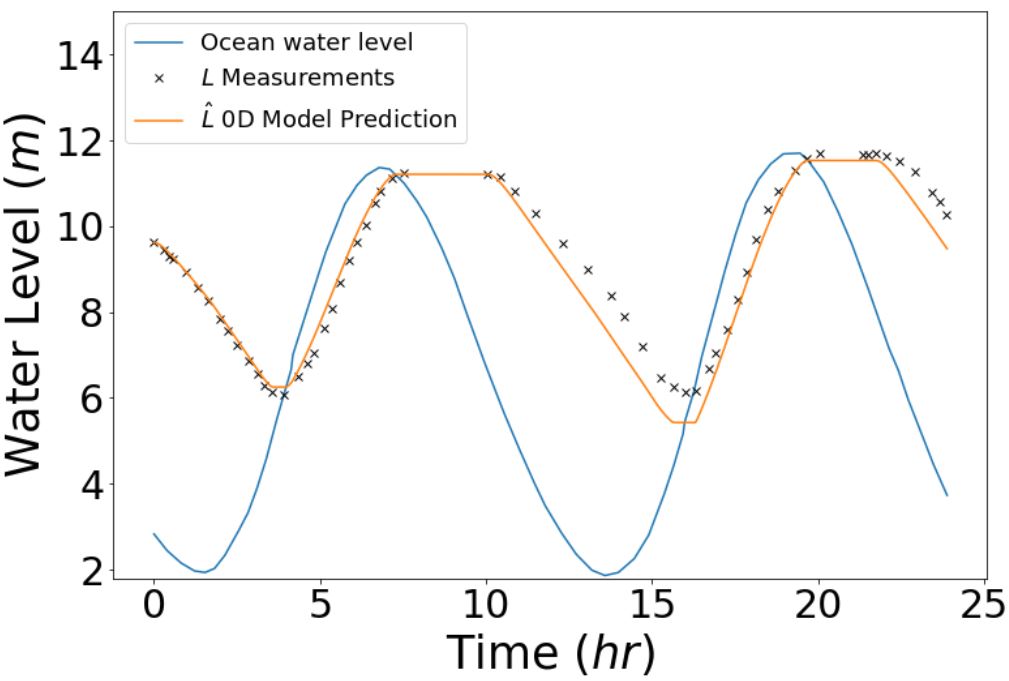}
    \caption{$\hat{L}$ vs $L$ comparison for $E_oG$.} \label{EoGVwl}
  \end{subfigure} \hfill \bigskip
  \begin{subfigure}[t]{.49\linewidth}
    \centering\includegraphics[width=\linewidth]{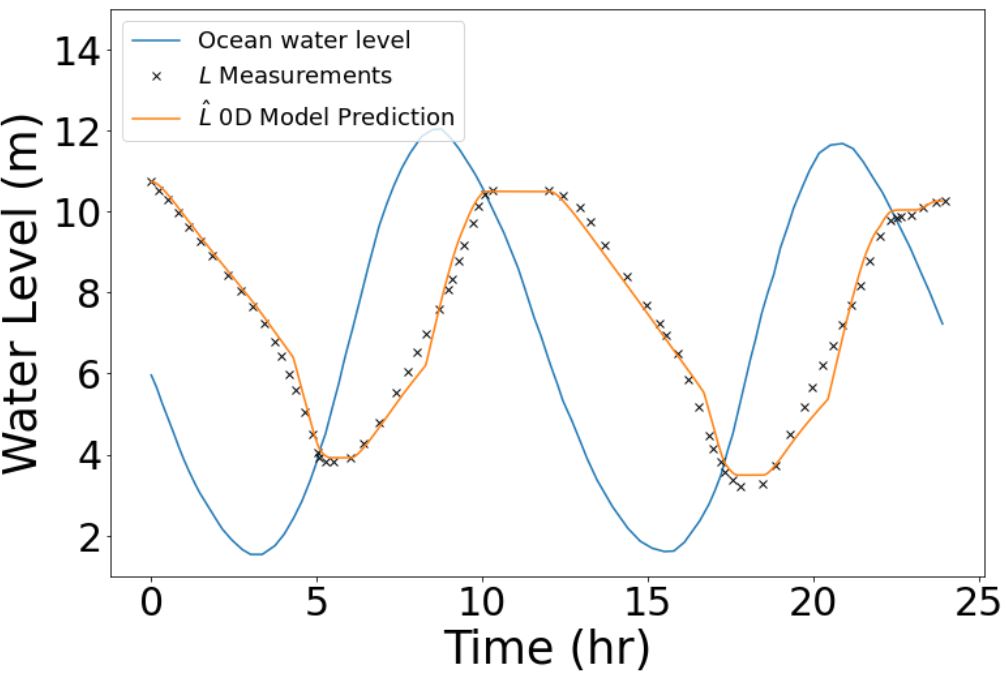}
    \caption{$\hat{L}$ vs $L$ comparison for $T.W.P$.} \label{TWPVwl}
  \end{subfigure} \hfill
  \begin{subfigure}[t]{.49\linewidth}
    \centering\includegraphics[width=\linewidth]{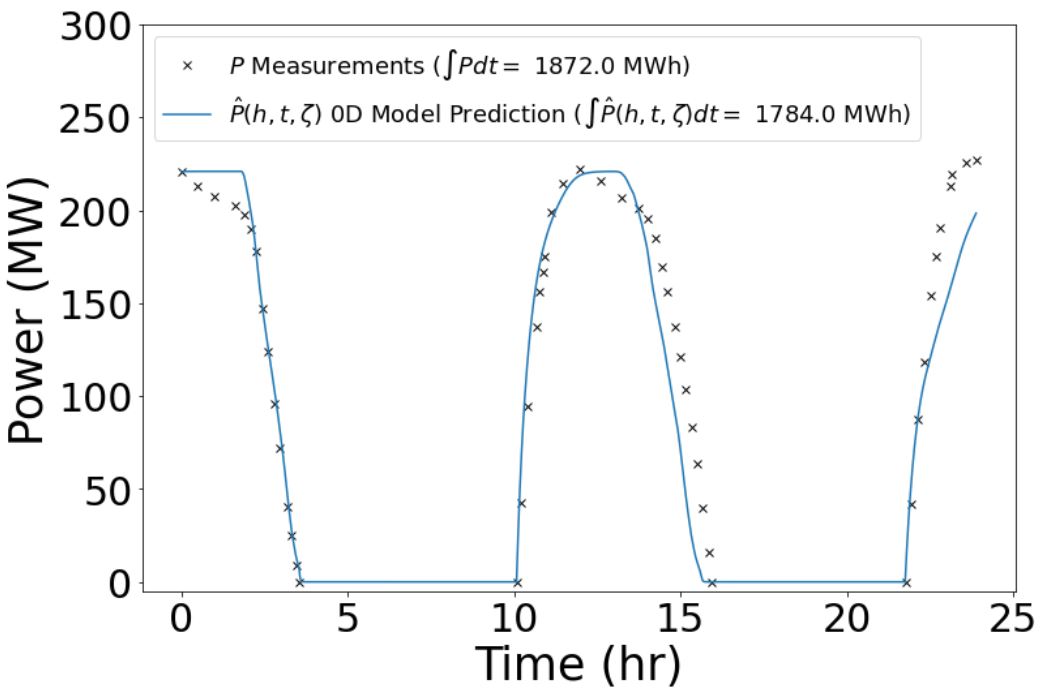}
    \caption{$\hat{P} (h, t, \zeta)$ vs $P$ comparison for $E_oG$.} \label{EoGVPow}
  \end{subfigure} \hfill \bigskip
  \begin{subfigure}[t]{.49\linewidth}
    \centering\includegraphics[width=\linewidth]{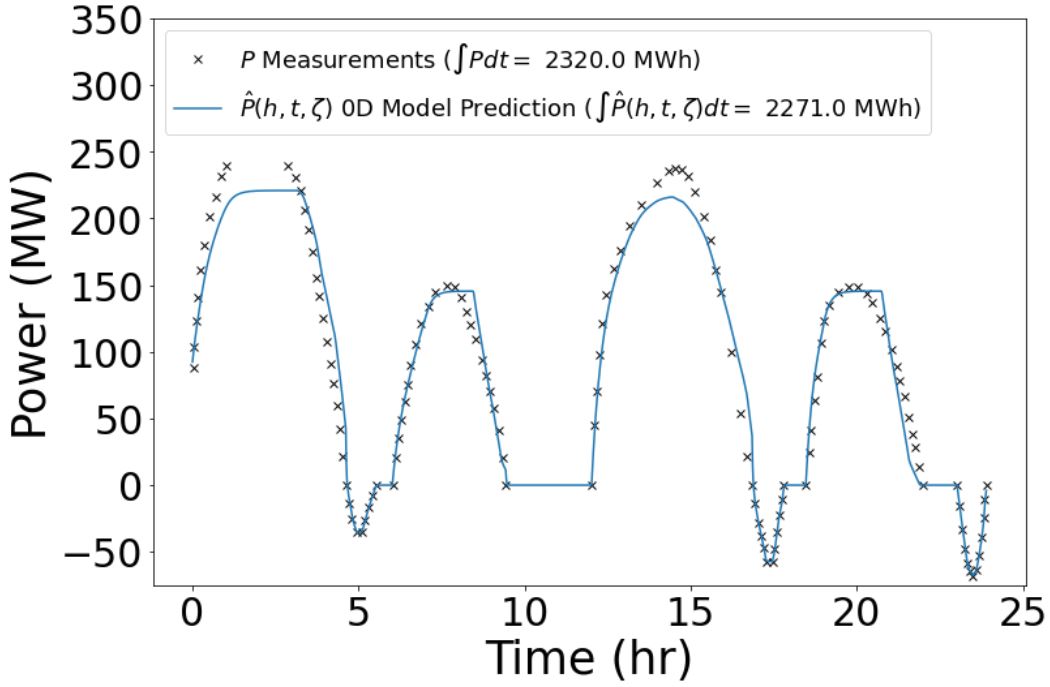}
    \caption{$\hat{P} (h, t, \zeta)$ vs $P$ comparison for $T.W.P$.}\label{TWPVPow}
  \end{subfigure}
  \begin{subfigure}[t]{.49\linewidth}
    \centering\includegraphics[width=\linewidth]{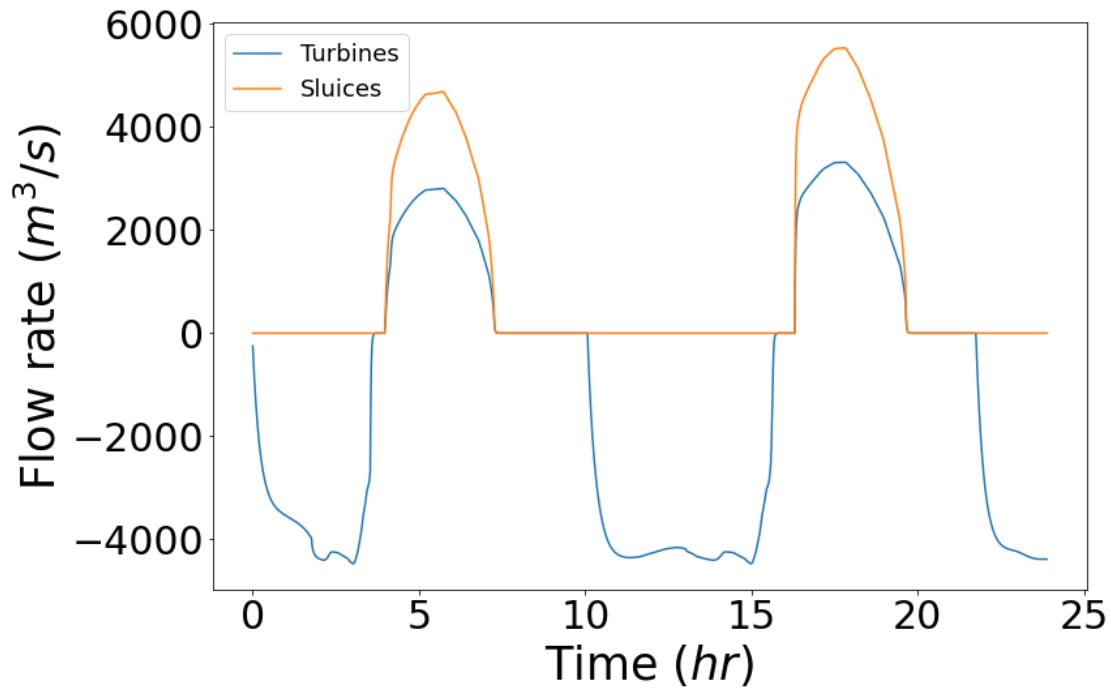}
    \caption{Turbines and sluices flow rates estimates for $E_oG$.} \label{EoGVQ}
  \end{subfigure}
  \begin{subfigure}[t]{.49\linewidth}
    \centering\includegraphics[width=\linewidth]{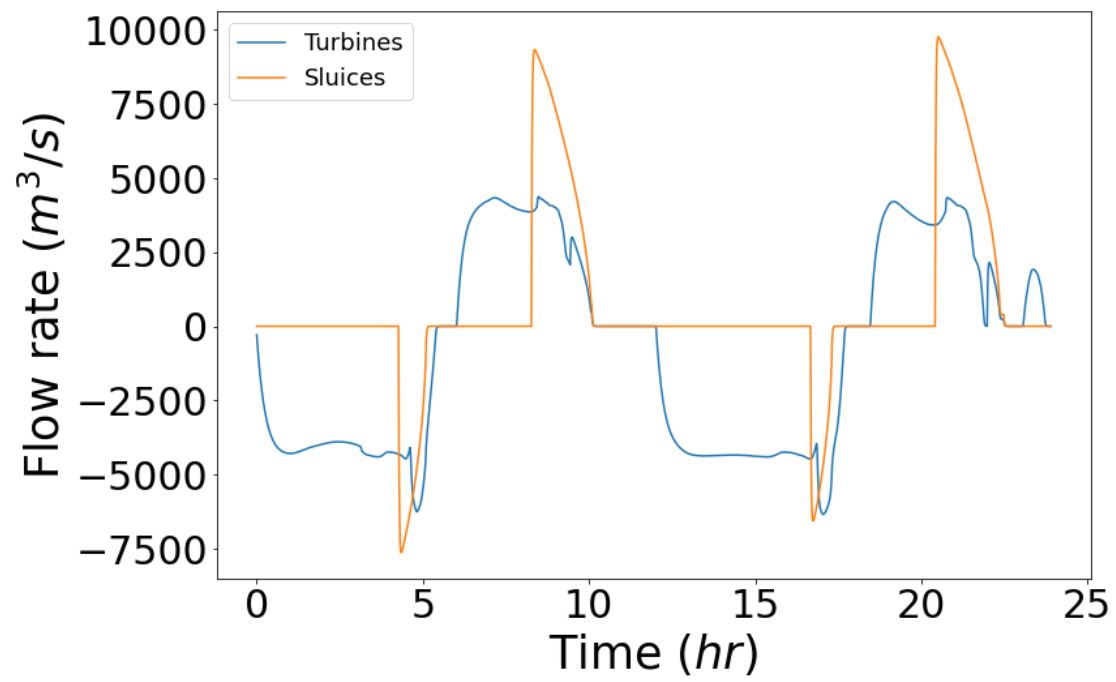}
    \caption{Turbines and sluices flow rate estimates for $T.W.P$.} \label{TWPVQ}
  \end{subfigure} \bigskip
  \caption{($a - d$) Validation of 0D La Rance model predictions (lagoon water levels and power generation) against site measurements from \cite{lebarbier1975power}. ($e, f$) Predicted turbines and sluices flow rates from 0D La Rance model.} \label{1stVerification}
\end{figure}

Utilising Fig. \ref{TWPRance} as reference, we set initial lagoon water levels $\hat{L}$ and turbine output power $\hat{P} (h, t, \zeta)$ to the same initial measured values ($L$ and $P$, respectively) at La Rance. Henceforth, we vary the operational mode for the hydraulic structures following the same timing as in Figs. \ref{EGRwl} and \ref{EGRPow}. When operating turbines in pump mode, the measured $P_{in}$ is applied to our pump model, so that pump flow rates can be predicted. A comparison of the 0D La Rance model predictions with measured data is shown in Fig. \ref{1stVerification}.

As noted in Section \ref{ParamDef}, sluice operation is expected at the end of power generation stages for the $T.W.P$ scheme (Figs. \ref{TWPVwl}, \ref{TWPVPow} and \ref{TWPVQ}). Since the timing for starting sluice operation was not provided by \cite{lebarbier1975power}, the showcased results for the $T.W.P$ scheme simulation assume a best fit between predicted and measured lagoon water levels. Nevertheless, the agreement between predicted and measured power outputs for both $E_oG$ and $T.W.P$ schemes is shown to be satisfactory. Indeed, by integrating predicted and measured power, energy deviation is only $4.7\%$ and $2.1\%$, for $E_oG$ and $T.W.P$ schemes comparisons, respectively.

While results in Fig. \ref{1stVerification} aim to validate the parametrisation techniques applied in reverse engineering La Rance into a 0D model, an optimal and comparable strategy for the control sequence of turbines and sluices is still required. In Section \ref{LaRanceDRL}, as a second validation step, we show that our trained DRL-Agent is capable of operating the 0D La Rance model with such strategy.

\section{Operational Optimisation of the AI-Driven La Rance Model}\label{LaRanceDRL}

In this section a DRL agent (modelled with Unity ML-Agents) is trained to operate the parametrised 0D model representation of the La Rance tidal barrage. We show that the obtained AI-Driven La Rance model is able to achieve an operational strategy that is comparable in (i) energy extraction capabilities and (ii) scheme of operation to the actual strategy utilised in La Rance. For our comparison analysis, we utilise measurements of the yearly net energy measured in \cite{sonnic2017rance} and the observed sequence of operation of hydraulic structures from \cite{lebarbier1975power}. It is important to highlight that the actual operation in La Rance, in contrast with the AI-Driven strategy, has the objective of maximising revenue instead of energy \cite{balls1988optimal}. However, as noted in \cite{harcourt2019utilising}, deviations in energy extraction when comparing revenue and energy based TRS optimisation are expected to be around $4-5\%$ only. Therefore, the comparisons we are showcasing are technically sound. In order to train and test our DRL agent, representative tidal data at the location of La Rance, are required. In the next section, we present a free software utilised for tidal prediction that can provide such data.

\subsection{JTides Training Data Validation}

JTides is a free, worldwide, tidal and current prediction software that utilises harmonic decomposition techniques for predicting ocean tides in several locations of the planet \cite{lutusJtides}. Although JTides has been utilised for research \cite{crockett2006tidal, nezlin2009dissolved}, its application in the field of tidal power has not yet been explored.

\begin{figure}[h]
  \centering
  \begin{subfigure}[t]{.37\linewidth}
    \centering\includegraphics[width=.95\linewidth]{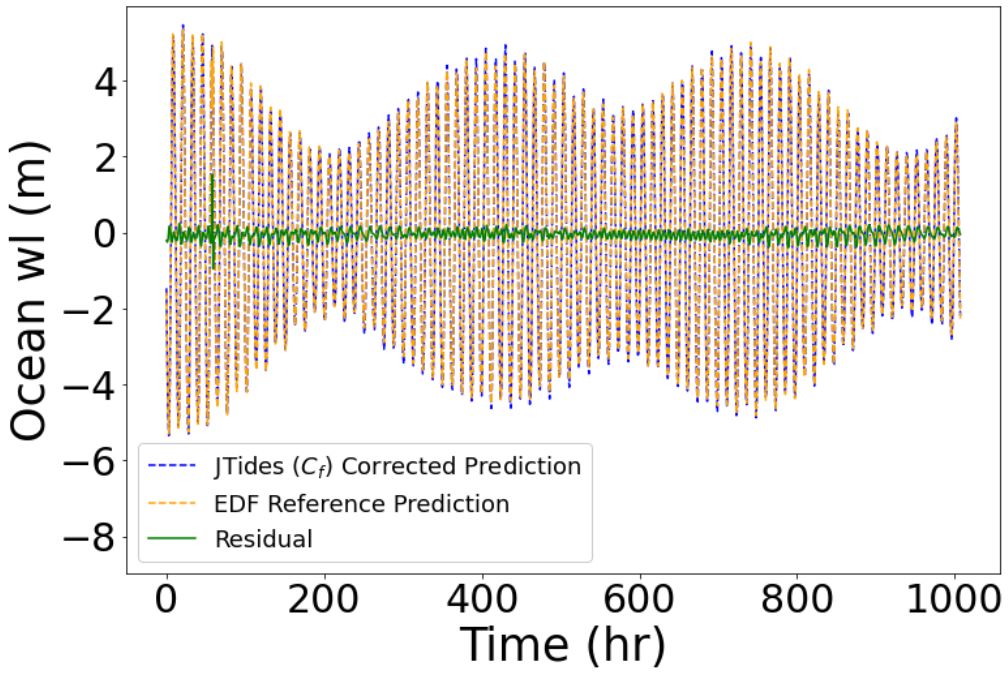}
    \caption{} \label{EDFJTides1}
  \end{subfigure} \hspace{10mm}
  \begin{subfigure}[t]{.37\linewidth}
    \centering\includegraphics[width=.95\linewidth]{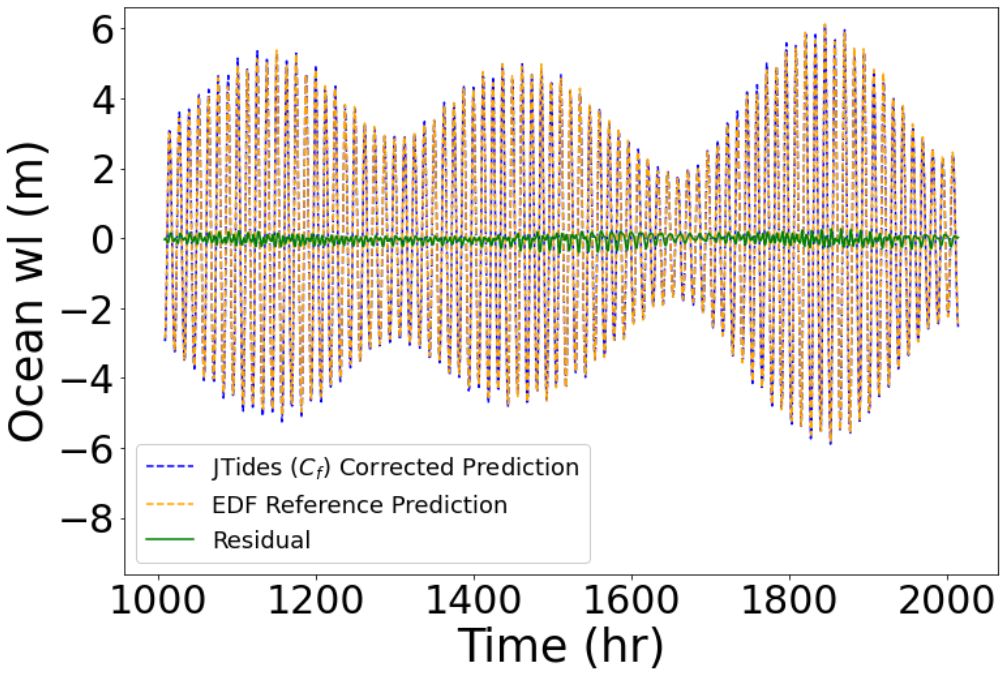}
    \caption{} \label{EDFJTides2}
  \end{subfigure} \hspace{10mm}
  \begin{subfigure}[t]{.37\linewidth}
    \centering\includegraphics[width=.95\linewidth]{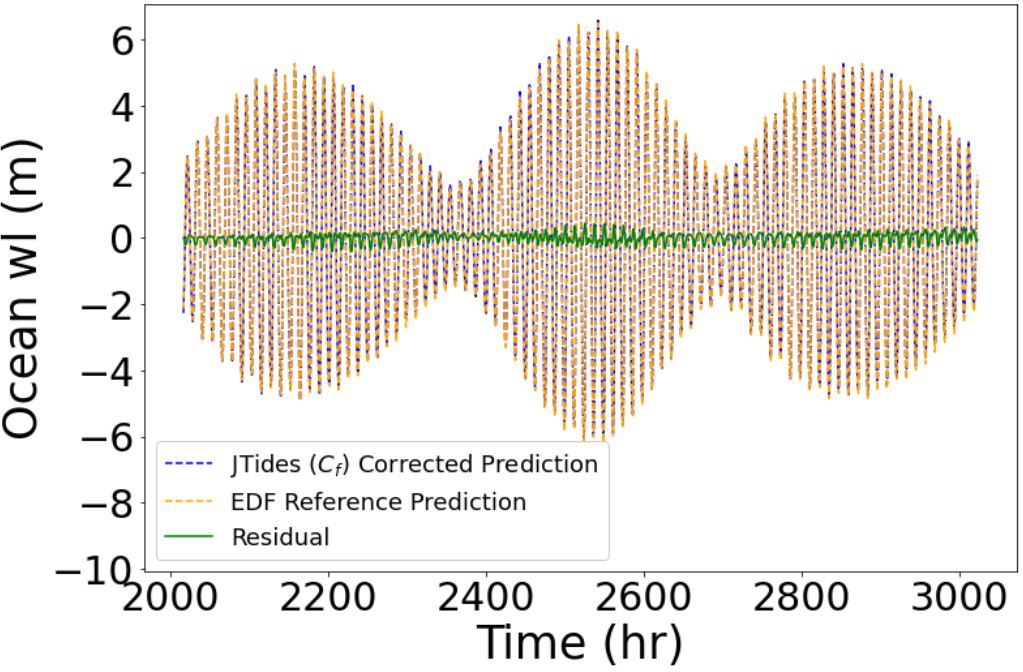}
    \caption{} \label{EDFJTides3}
  \end{subfigure} \hspace{10mm}
  \begin{subfigure}[t]{.37\linewidth}
    \centering\includegraphics[width=.95\linewidth]{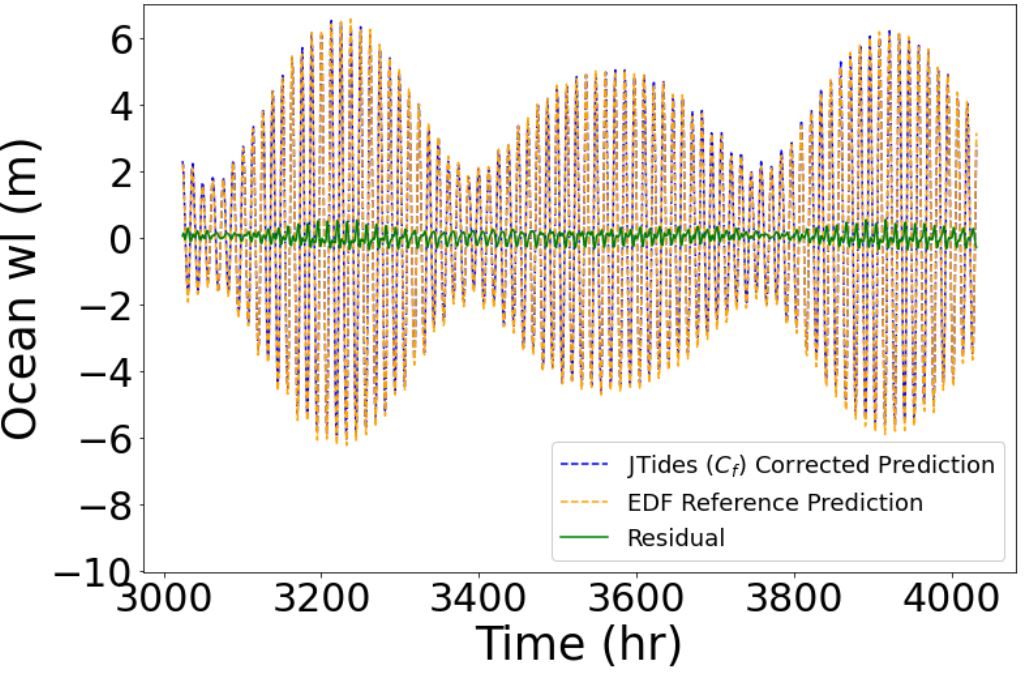}
    \caption{} \label{EDFJTides4}
  \end{subfigure} \hspace{10mm}
  \begin{subfigure}[t]{.37\linewidth}
    \centering\includegraphics[width=.95\linewidth]{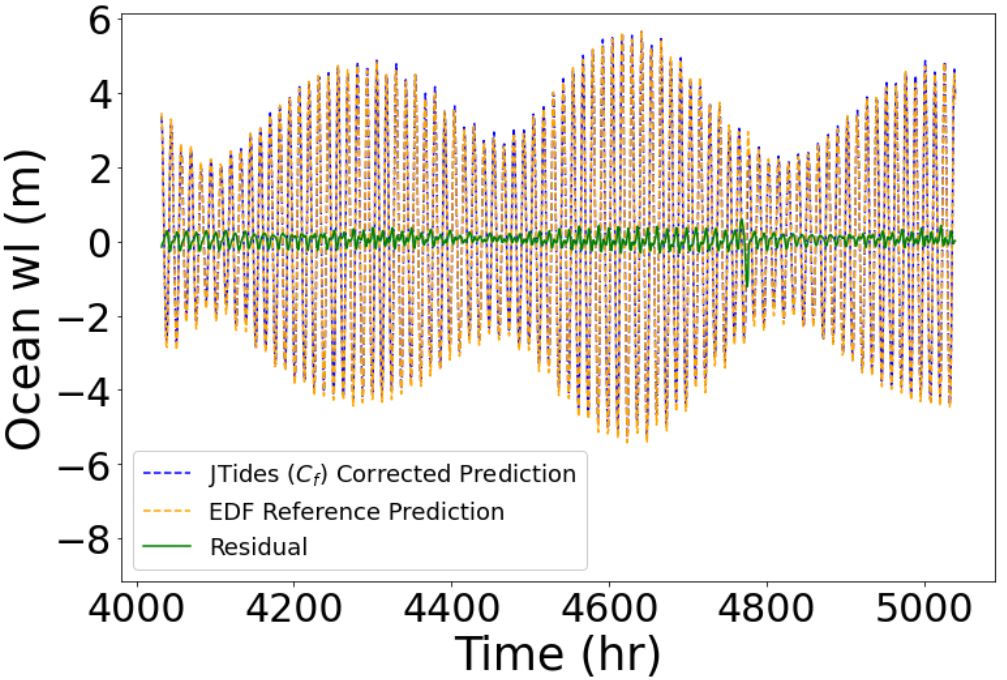}
    \caption{} \label{EDFJTides5}
  \end{subfigure} \hspace{10mm}
  \begin{subfigure}[t]{.37\linewidth}
    \centering\includegraphics[width=.95\linewidth]{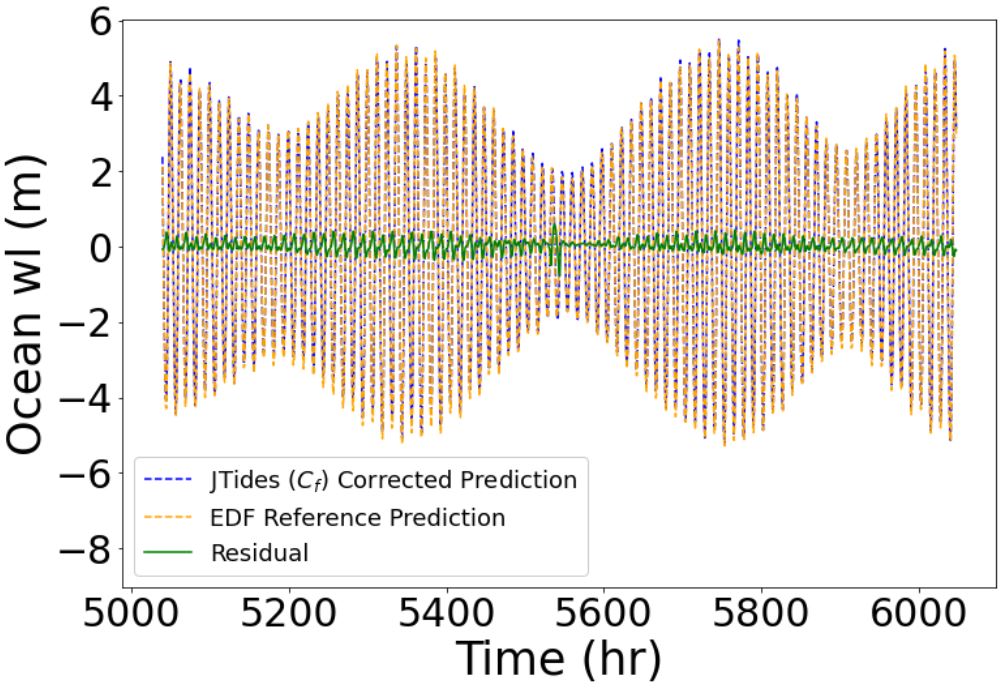}
    \caption{} \label{EDFJTides6}
  \end{subfigure} \hspace{10mm}
  \begin{subfigure}[t]{.37\linewidth}
    \centering\includegraphics[width=.95\linewidth]{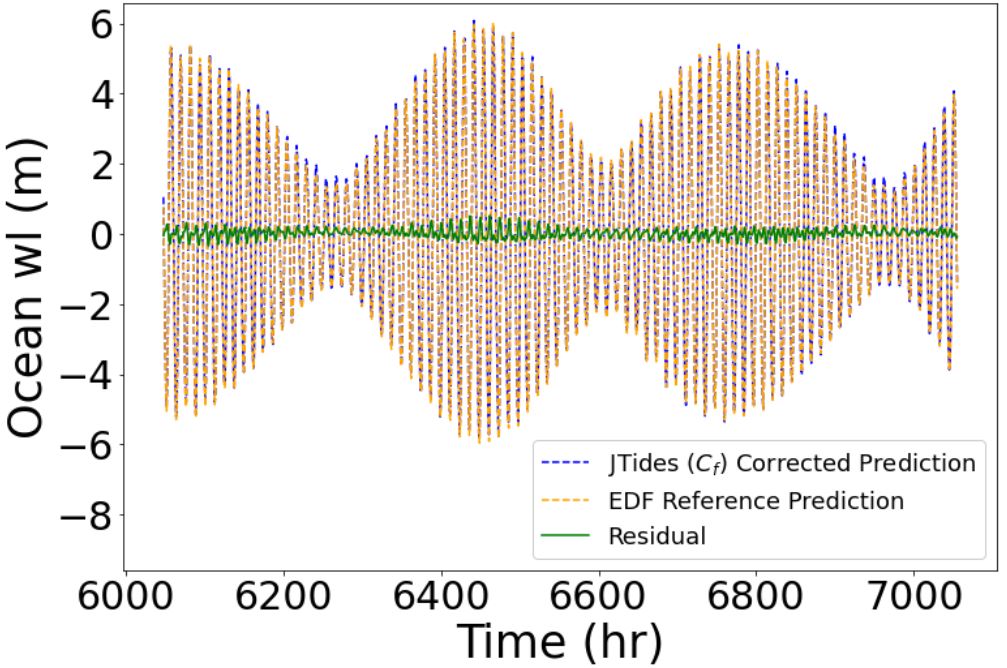}
    \caption{} \label{EDFJTides7}
  \end{subfigure} \hspace{10mm}
  \begin{subfigure}[t]{.37\linewidth}
    \centering\includegraphics[width=.95\linewidth]{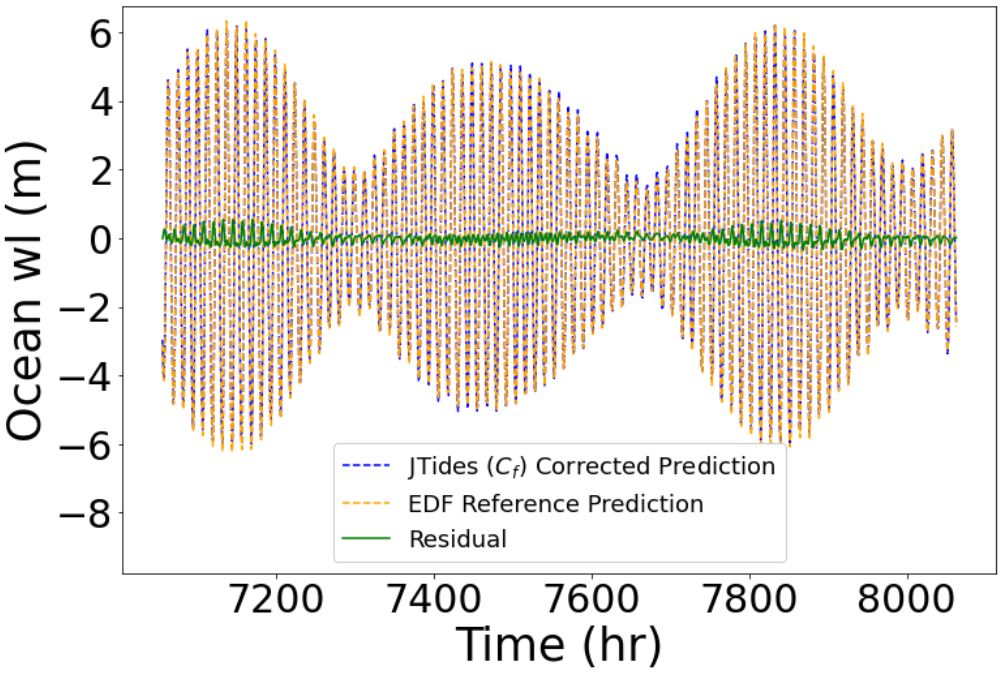}
    \caption{} \label{EDFJTides8}
  \end{subfigure}
  \caption{Comparisons between $C_f$ corrected JTides and EDF ocean predictions, for the La Rance tidal barrage, for one year-round ($June/06/2020$ up to $June/05/2021$).} \label{EDFJTidesComp}
 \end{figure}

From JTides interface, the user can insert coordinates for any location on Earth. From this location, JTides looks for a nearest point of reference, for which it can extract tidal predictions from its database. By providing the coordinates of the La Rance tidal barrage, JTides returns tidal predictions for the location of St. Helier, Jersey, the largest of the Channel Islands in the English Channel. This island is located around $80~km$ from La Rance and, as the latter, experiences one of the highest tidal ranges on the planet (up to $10~m$ \cite{cooper2005sediment}). In order to assess if tidal predictions from St. Helier are appropriate for La Rance, tidal predictions provided by EDF, the company responsible for operating La Rance, are utilised as reference. The tidal predictions from EDF were available through the web-page (https://www.edf.fr/usine-maremotrice-rance/marees-en-rance), being updated every week, providing $3$ weeks of forecast. A whole year of tidal predictions were collected manually, for comparison with JTides predictions. By comparing JTides and EDF predictions, we note that results have the same pattern, although with a small deviation at tidal range's extremes, with tidal predictions by EDF consistently predicting higher tidal amplitudes. By assuming that EDF and JTide's waves are similar, a deviation coefficient between tidal predictions can be obtained through the simple method of root mean square differences (RMS) \cite{mcnatt2020comparison}. With this assumption, we can estimate a correction factor $C_f$ to be applied to JTides's prediction ($JTides \times C_f$), in order to reduce the deviation between JTides and EDF estimates:
\begin{equation} \label{CorrectionFactor}
    C_f = \sum_{i = 1}^{i = N} O_{Ji} O_{Ei}/O_{Ji}^2,
\end{equation}
where $O_{Ji}$ and $O_{Ei}$ are the ocean predictions for JTides and EDF (oscillating around mean water level), respectively, and $N$ is the number of data points for a whole year. With this method, a $C_f \approx 1.10$ was obtained. A comparison of tidal prediction elevations from EDF and $JTides \times C_f$ is showcased in Fig. \ref{EDFJTidesComp}, for one year round. From the observed residuals, we see that the agreement between $JTides \times C_f$ and EDF is consistent throughout the year, apart from very few isolated spikes. Therefore, we resume our work with the assumption that $JTides \times C_f$ tidal predictions are representative of expected ocean water levels at La Rance. With this assumption, we utilise the $JTides \times C_f$ software capabilities to generate training data (from $2013$ until $2038$) for our DRL agent.

\subsection{Agent-Environment Setup}

\subsubsection{Unity 3D \& Unity ML-Agents}

The Unity3D graphics engine is a popular game developing environment that has been used to create games and simulations in 2D and 3D since its debut in 2005. It has received
widespread adoption in other areas as well, such as architecture, engineering and construction \cite{juliani2018unity}.

Unity ML-Agents is an open-source project that allows for designing environments where a smart agent can learn through interactions \cite{juliani2018unity}. It has been chosen in this project due to ease of implementation, built-in PPO algorithm and visual framework for visualising real-time control of TRS.

\subsubsection{Agent-Environment MDP Modelling and Training} \label{MDP}

To formalise La Rance operation as a DRL problem (and solve the DRL problem through PPO) we need to design an MDP in Unity ML-Agents with environment, agent, actions, states and reward components (Section \ref{DRLback}).

\begin{figure}[h]
  \centering
  \begin{minipage}[b]{.5\textwidth}
    \includegraphics[width=1\linewidth]{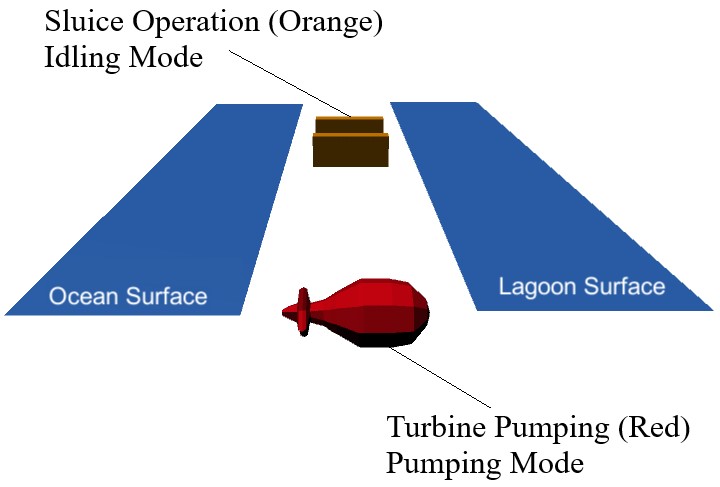}
  \end{minipage}
  \caption{Unity ML-Agents MDP environment for a 0D model of the La Rance tidal barrage during $E.P$.} \label{EbbPumpUnity}
\end{figure}

For the environment component (as in \cite{moreira2021prediction}), simple representative 3D models for turbines, sluices, ocean and lagoon are created in Unity3D and then imported to a Unity ML-Agents project. Together, these 3D elements visually compose the training environment for our MDP. In this environment, the equations for simulating turbines (in power generation, pumping and idling modes), sluices and equivalent lagoon wetted area are extracted from Section \ref{ParamDef}. For both training and test stages $JTides \times C_f$ outputs are used for our ocean component, while lagoon water level motion is dictated by Eq.~\ref{Num0D}. The developed 3D representations of turbines and sluices change colours according to the operational mode chosen by the agent. For the turbine, green represents power generation mode, orange -- idling mode, black -- offline mode (zero flow rate) and red -- pumping mode. Similarly, sluices change colour between orange and black for sluicing and offline modes, respectively. Fig. \ref{EbbPumpUnity} shows a capture of the Unity ML-Agents MDP environment representation of the parametrised 0D model of the La Rance tidal barrage during $E.P$, with the representative models for sluice and turbines in idling and pump generation modes, respectively. Ocean and Lagoon surface level motion are also represented.

\begin{figure}[h]
  \centering
  \begin{minipage}[b]{.5\textwidth}
    \includegraphics[width=1\linewidth]{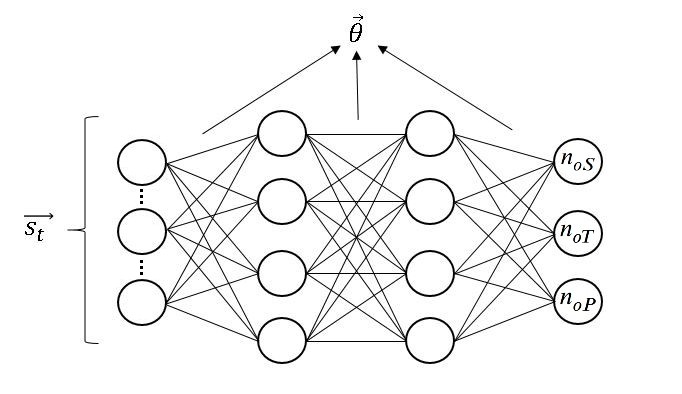}
  \end{minipage}
  \caption{Input-output representation of policy (actor) neural network with $3$ branches of discrete actions.} \label{NeuralNetworkPolicy}
\end{figure}

As in \cite{moreira2021prediction} the DRL agent (actor neural network) is responsible for controlling turbine and sluice operational modes through node outputs $n_{o}$, according to a vector of input states $s_t$. In this case study however, a discrete control solution was adopted when utilising the PPO algorithm, which allowed for the agent to learn a policy that uses turbines in pump mode -- in contrast with continuous control solution that ignored turbine pumping capabilities completely. The control solution outputs $3$ branches in the last layer of the policy network. The first branch $n_{oS}$ is responsible for controlling sluices, while the second branch $n_{oT}$ is responsible for setting the turbine operational mode. Finally, the third branch $n_{oP}$ controls the input power $P_{in}$ available to turbines in pump mode (Fig. \ref{NeuralNetworkPolicy}). Each branch ramifies into possible discrete actions that can be taken (i.e. a probability mass function), allowing for the full range of combinations between turbines and sluices to be explored during training. The possible actions for each branch are showcased in Table \ref{DiscreteLarance}, while the required input states for the policy neural network are presented in Table \ref{StatesLaRance}. 


\begin{table}[width=.5\linewidth,cols=2,pos=h]
\centering \footnotesize
  \caption{Discrete control options for La Rance.}\label{DiscreteLarance}
\begin{tabular*}{\tblwidth}{@{} LLLL@{} }
\toprule
Branch       & Discrete Action  \\ \hline
$n_{oS}$ (sluices)  & Offline           \\
             & Online          \\ \hline
             & Offline   \\
             & Ebb Generation \\
$n_{oT}$ (turbines) & Flood Generation          \\
             & Idling           \\
             & Pumping          \\ \hline
$n_{oP}$ (pump $P_{in}$)  & $\{P_{in}: P_{in} = 0.25 MW \times n_{oP}, n_{oP} \in \{0, 1, ... 16\}\}  $ \\
  \bottomrule
\end{tabular*}
\end{table}

\begin{table}[width=.5\linewidth,cols=2,pos=h]
\centering \footnotesize
  \caption{Input states for PPO neural network, for La Rance.}\label{StatesLaRance}
  \begin{tabular*}{\tblwidth}{@{} LLLL@{} }
    \toprule
	States (at times $t$ and $t-1$)  & Units  \\
    \midrule
	Ocean water level & ``Normalised'' $[0,1]$ (float)\\
	Lagoon water level & ``Normalised'' $[0,1]$ (float) \\
	Sluice Mode &  $n_{oS} \in \{0, 1\}$ (integer) \\
	Turbine Mode & $n_{oT} \in \{0, 1, 2, 3, 4\}$ (integer) \\
	Pump Power Input &  $n_{oP} \in \{0, 1, ... 16\}$ (integer) \\
  \bottomrule
\end{tabular*}
\end{table}


As shown in Table \ref{DiscreteLarance}, $P_{in}$ values have been discretised, with an upper bound of $4 MW$, for each bulb unit. This upper bound has been selected, given that initial training sessions for La Rance never surpassed $4 MW$ power input, for each of the $24$ bulb units available (Table \ref{LaRanceDesign}). Furthermore, as discussed in \cite{lebarbier1975power}, even though each unit can receive $6 MW$ of power input, optimal results were obtained with smaller values, with pumping sometimes being restricted to $50 MW$ (for all $24$ units), depending on tidal range \cite{lebarbier1975power}.

\begin{figure}[h]
	\centering
	\includegraphics[width=.5\linewidth]{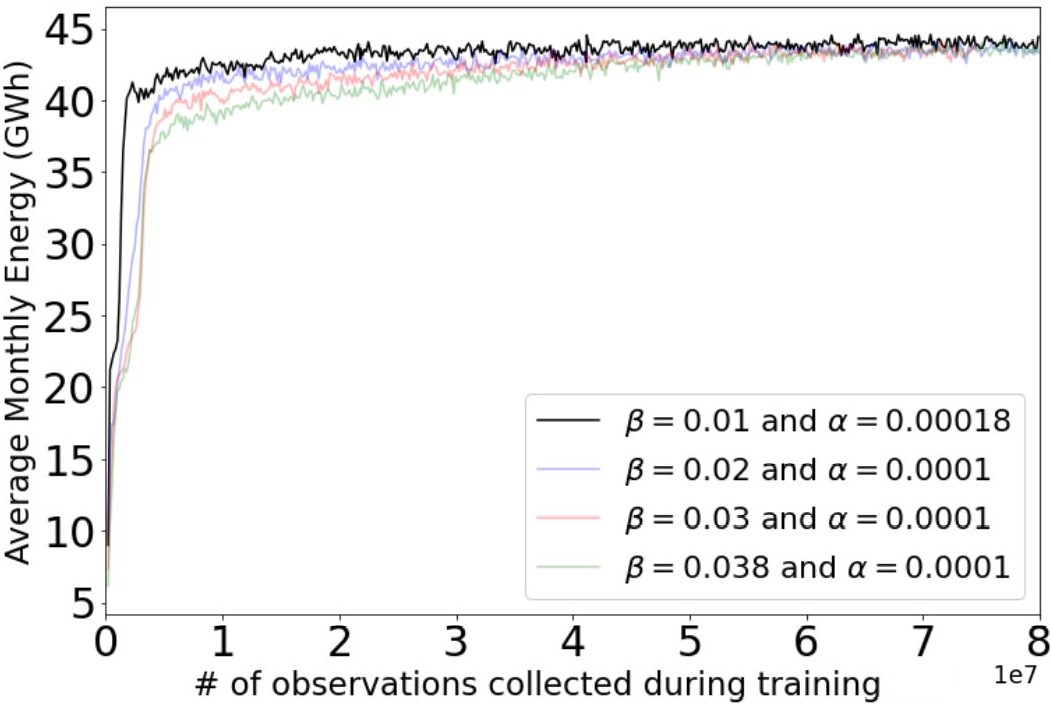}
	\caption{Monthly cumulative energy (reward) in GWh, averaged for all 64 environments during parallel training, for the AI-Driven La Rance model.}
	\label{LaRanceTraining} 
\end{figure}

Parallel training is performed with 64 copies of the environment in Unity and observing the average monthly energy reward obtained. By starting with the initial hyperparameter settings from the optimised model in \cite{moreira2021prediction} as reference, we note a significant improvement of results when increasing the number of units and hidden layers of the policy neural network. After fixing the complexity of the neural network, converging to a stable plateau was possible by tuning hyperparameters $\alpha$ and $\beta$ for multiple runs (Fig. \ref{LaRanceTraining}), with $\alpha$ being the learning rate and $\beta$ a linear decaying hyperparameter responsible for controlling how much the agent explores the environment during training (i.e., increasing $\beta$ leads to more random actions at the beginning of training) \cite{moreira2021prediction}. From initial $\beta = 0.038$ and $\alpha = 0.0001$, an optimal policy was obtained by decreasing $\beta$ and increasing $\alpha$. The final hyperparameter setting for acquiring the optimal policy and the utilised Unity version are are shown in the Supplementary Material.

After training, the DRL agent performs real-time optimal control of the hydraulic structures, without the need of future tidal inputs (in contrast with state-of-art methods) \cite{moreira2021prediction}.




\subsection{Agent Performance Evaluation}
\begin{figure}[h]
	\centering
	\includegraphics[width=.5\linewidth]{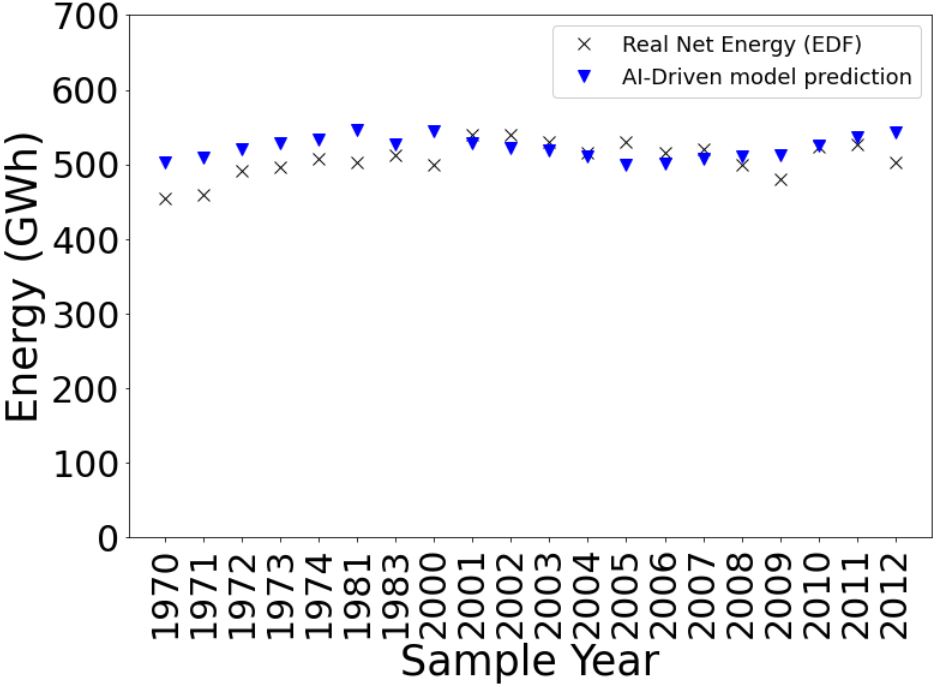}	\caption{Comparison between measured and predicted yearly energy outputs for the La Rance Barrage.}
	\label{LaRanceTesting} 
\end{figure}

With our trained DRL agent, we proceed to compare the yearly energy extraction capabilities of our real-time AI-Driven model against measured data by EDF \cite{sonnic2017rance}. In order to do so, tidal predictions for the reference years in \cite{sonnic2017rance} are produced with $JTides \times C_f$ and used as inputs for the AI-Driven La Rance model. A comparison between the predicted and measured yearly energy generated by La Rance is shown in Fig. \ref{LaRanceTesting}, where we can see a satisfactory agreement of results, with La Rance's AI-Driven predictions and EDF measurements averaging $521.5 MW$ and $507.4 MW$, respectively. The $2.6\%$ average gain from the DRL operation is expected, given that energy oriented optimisation schemes have been shown to attain up to $5\%$ more energy than revenue oriented optimisation strategies \cite{harcourt2019utilising}.

\begin{figure}[h]
  \centering
  \begin{subfigure}[t]{.47\linewidth}
    \centering\includegraphics[width=\linewidth]{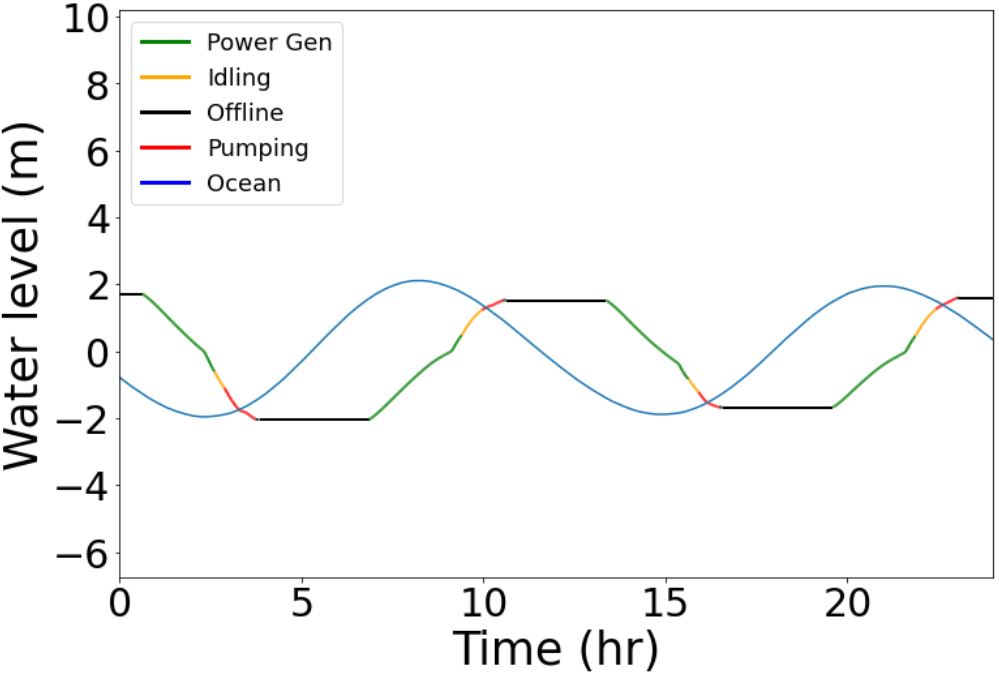}
    \caption{Ocean and lagoon water level variations.} \label{LaRanceTR1}
  \end{subfigure}\hfill
  \begin{subfigure}[t]{.47\linewidth}
    \centering\includegraphics[width=\linewidth]{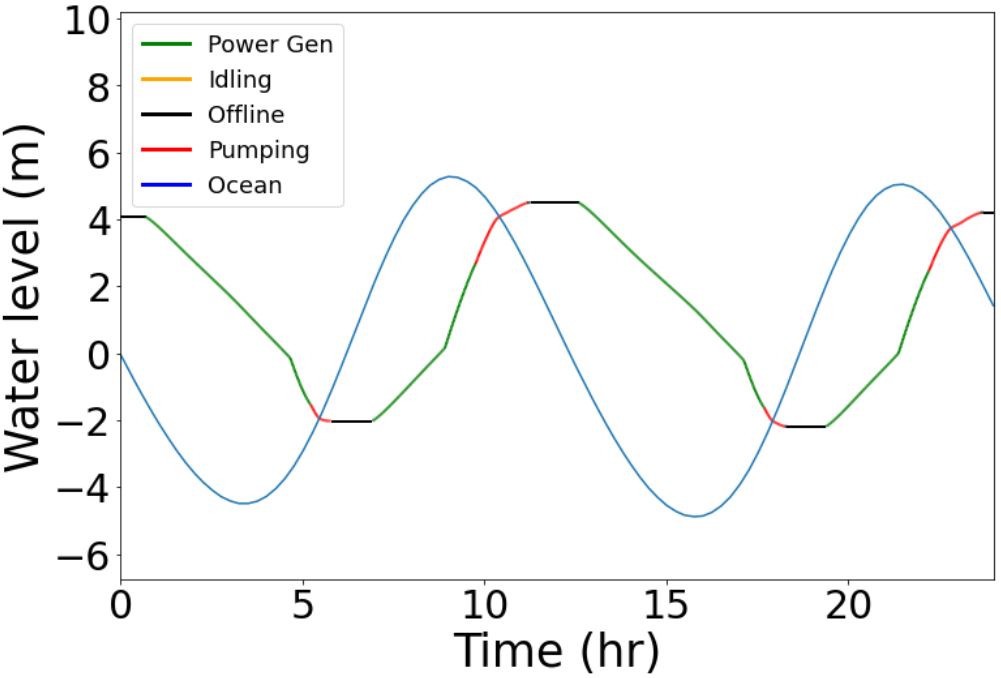}
    \caption{Ocean and lagoon water level variations.} \label{LaRanceTR2}
  \end{subfigure}
  \begin{subfigure}[t]{.47\linewidth}
    \centering\includegraphics[width=\linewidth]{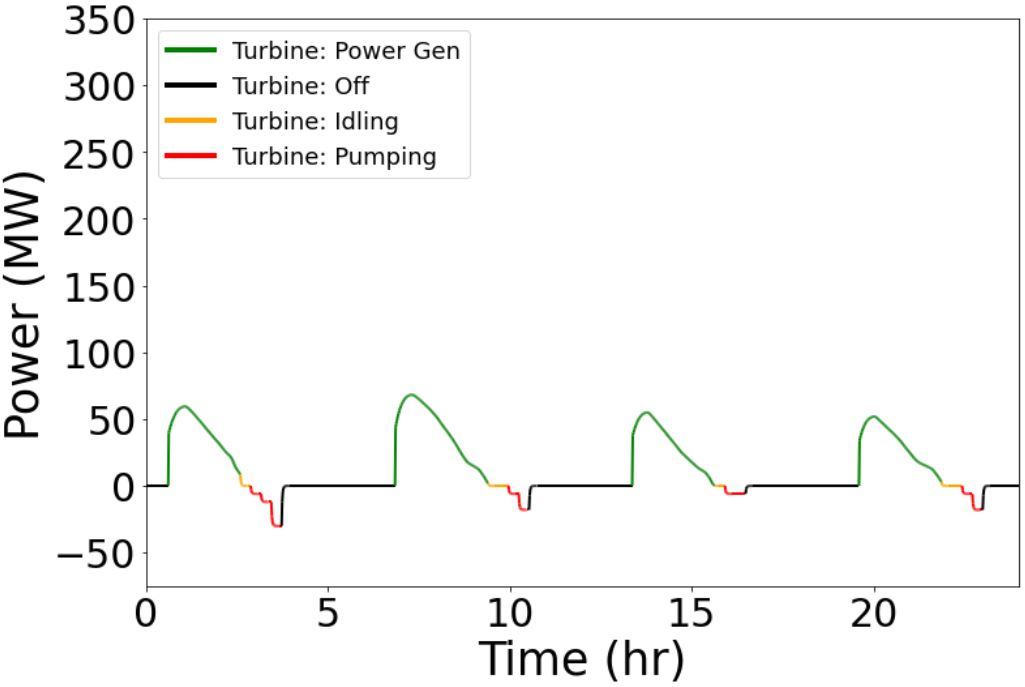}
    \caption{Power output.} \label{LaRancePow1}
  \end{subfigure}\hfill
  \begin{subfigure}[t]{.47\linewidth}
    \centering\includegraphics[width=\linewidth]{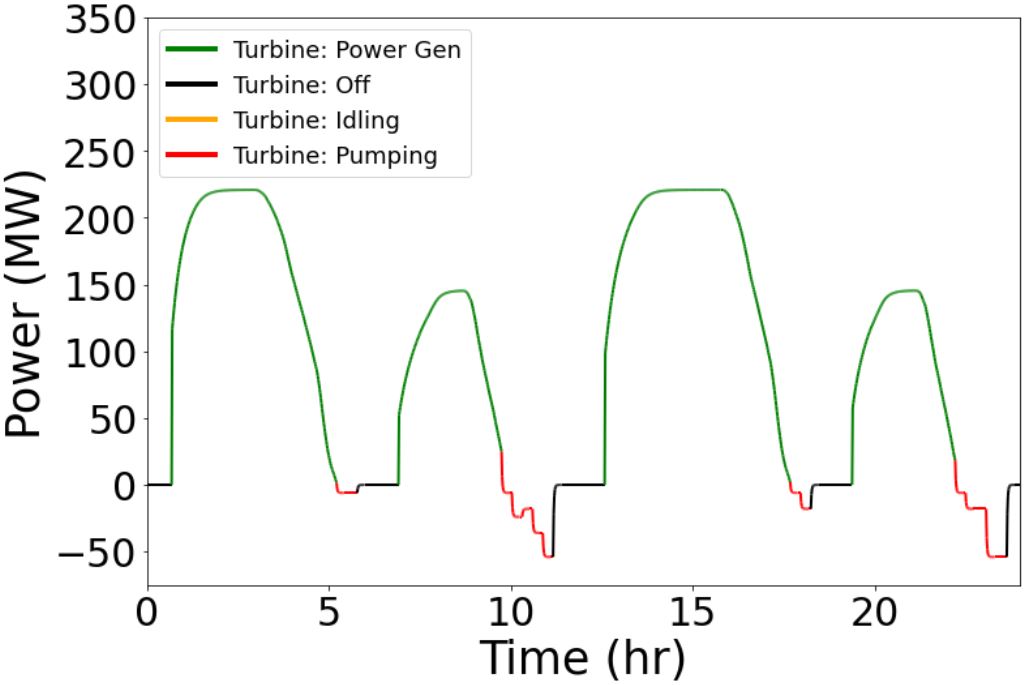}
    \caption{Power output.} \label{LaRancePow2}
  \end{subfigure}
  \begin{subfigure}[t]{.47\linewidth}
    \centering\includegraphics[width=\linewidth]{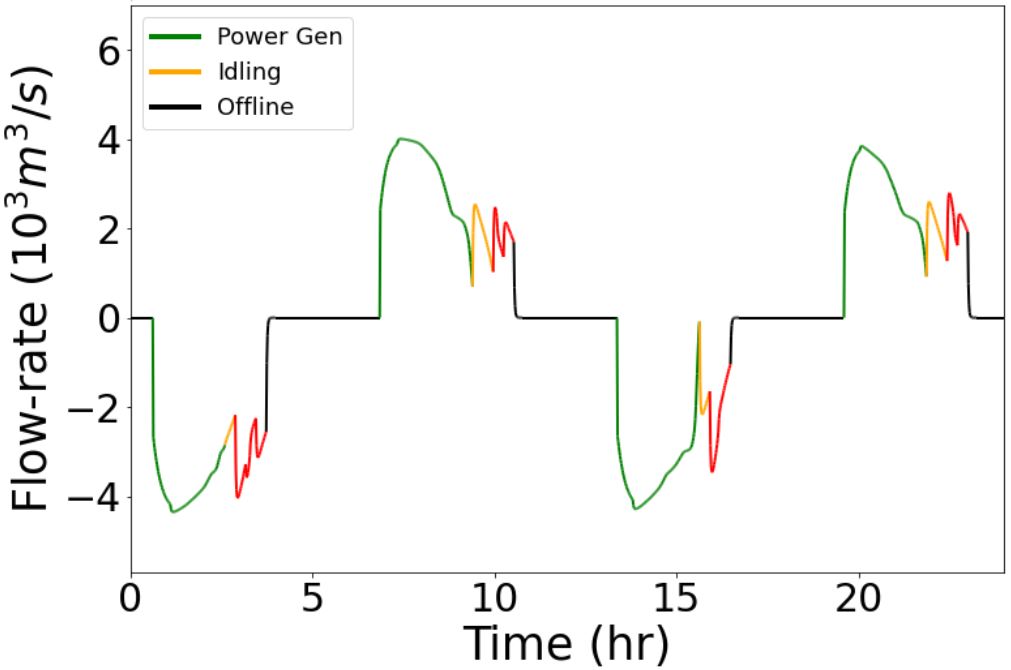}
    \caption{Turbine flow rate.} \label{LaRanceQt1}
  \end{subfigure}\hfill
  \begin{subfigure}[t]{.47\linewidth}
    \centering\includegraphics[width=\linewidth]{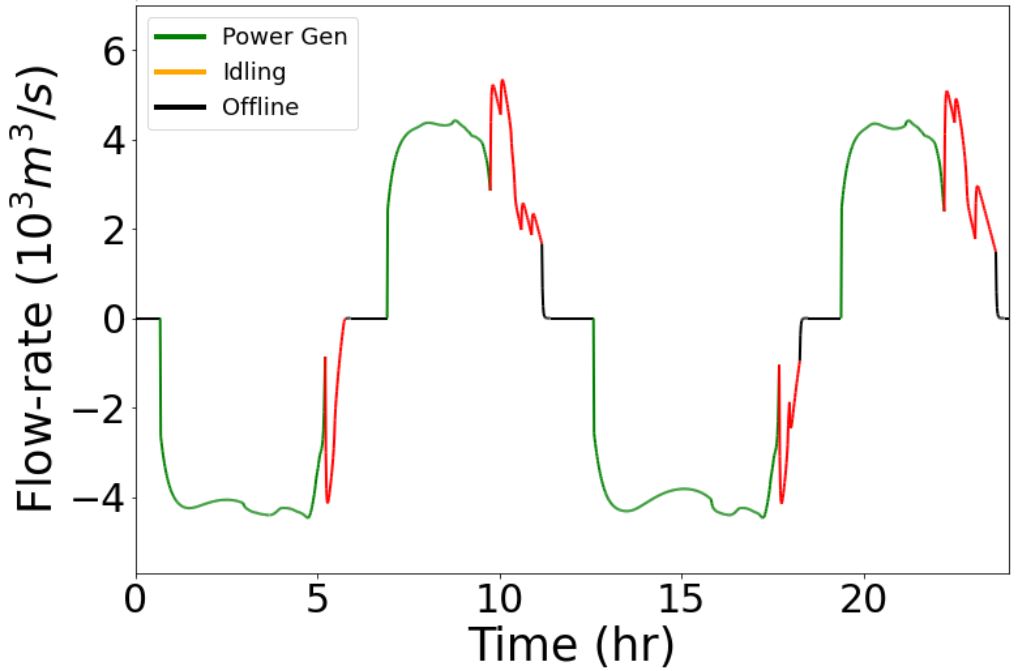}
    \caption{Turbine flow rate.} \label{LaRanceQt2}
  \end{subfigure}
  \caption{Lagoon water level variations, power output and predicted flow rate coloured according to turbine operation from the AI-Driven La Rance model, for small ($a, c, e$) and large ($b, d, f$) tidal ranges. In ($a, b$), ocean is coloured in blue.} \label{LaRanceTidalRange}
\end{figure}

From the collected data provided by the AI-Driven model, we observe that the resulting optimal policy chooses to operate La Rance with a $T.W.P$ scheme strategy, independently of tidal range (Fig. \ref{LaRanceTidalRange}). The obtained optimal policy consistently starts turbine pump mode when a positive pump head is still available (Figs. \ref{LaRanceTR1} and \ref{LaRanceTR2}), which is (i) not possible for state-of-art methods and (ii) the same strategy adopted in the actual operation of La Rance (Fig. \ref{EGRPow}). Also exclusive to the AI-Driven model is its capability of fine-tuning power input for turbines in pump mode (Figs. \ref{LaRancePow1} and \ref{LaRancePow2}), in a similar manner to what is observed in La Rance (Fig. \ref{EGRPow}). Furthermore, we note that for tidal ranges above $\approx 3~m$, turbine idling mode is not utilised by the agent (i.e. turbine in power generation mode directly switch to pumping mode), indicating the DRL agent capability of adjusting its strategy according to the observed tidal range. 

\begin{figure}[h]
  \centering
  \begin{subfigure}[t]{.45\linewidth}
    \centering\includegraphics[width=\linewidth]{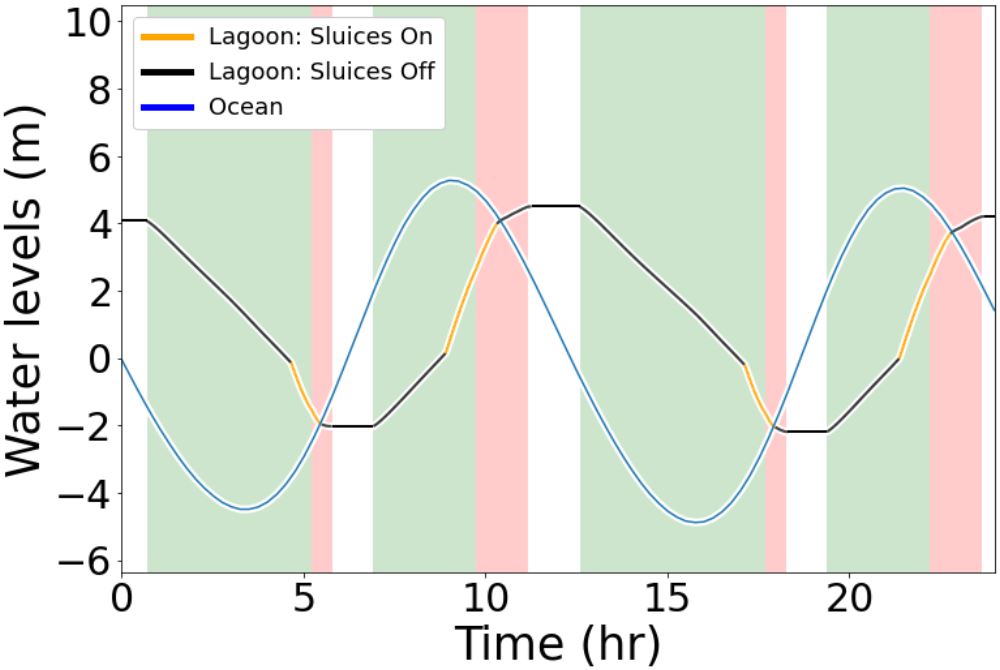}
    \caption{Ocean and lagoon water level variations.} \label{LaRanceSlWl}
  \end{subfigure}\hfill
  \begin{subfigure}[t]{.47\linewidth}
    \centering\includegraphics[width=\linewidth]{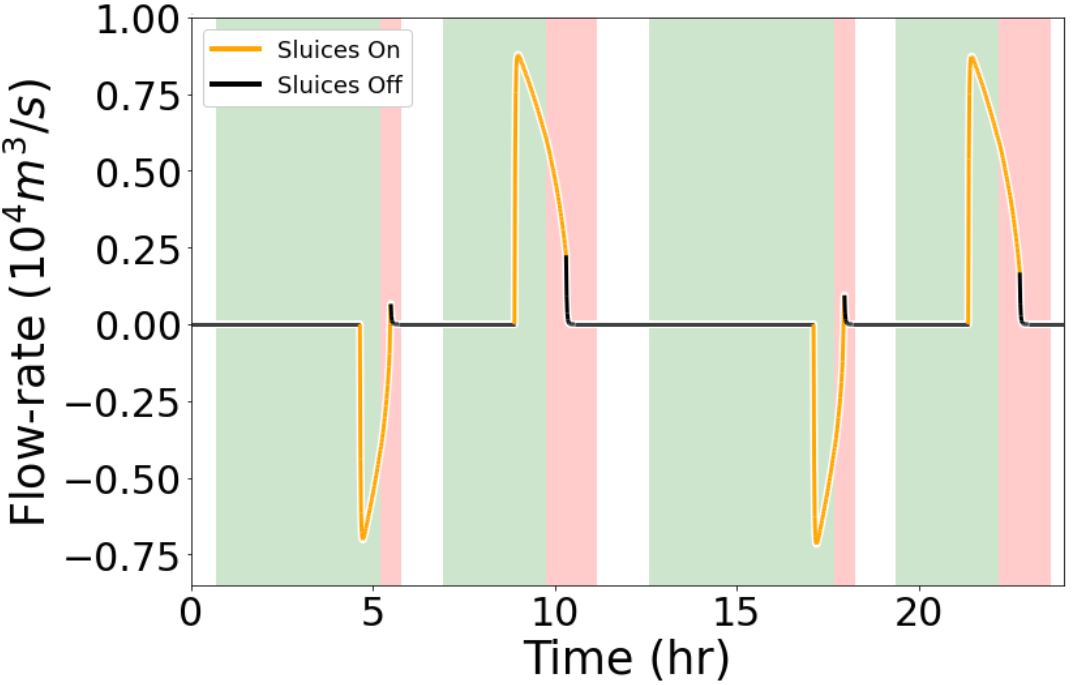}
    \caption{Sluice flow rate.}
    \label{LaRanceSlQ}
  \end{subfigure}
  \caption{Lagoon water level variations and predicted flow rate coloured according to sluice operation from the AI-Driven La Rance model. In ($a$), ocean is coloured in blue. In both images, background is coloured following turbine operation, with green and red representing power generation and pumping modes, respectively.} \label{sluiceDRLLaRance}
\end{figure}

Finally, by observing sluice operation in Fig. \ref{sluiceDRLLaRance}, we note that sluices are operated independently from turbines, characterising the variant operation of TRS (shown to be superior to the classical operation of TRS when the goal is maximising power generation \cite{moreira2021prediction}).

\begin{figure}[h!]
  \centering
  \begin{subfigure}[t]{.45\linewidth}
    \centering\includegraphics[width=\linewidth]{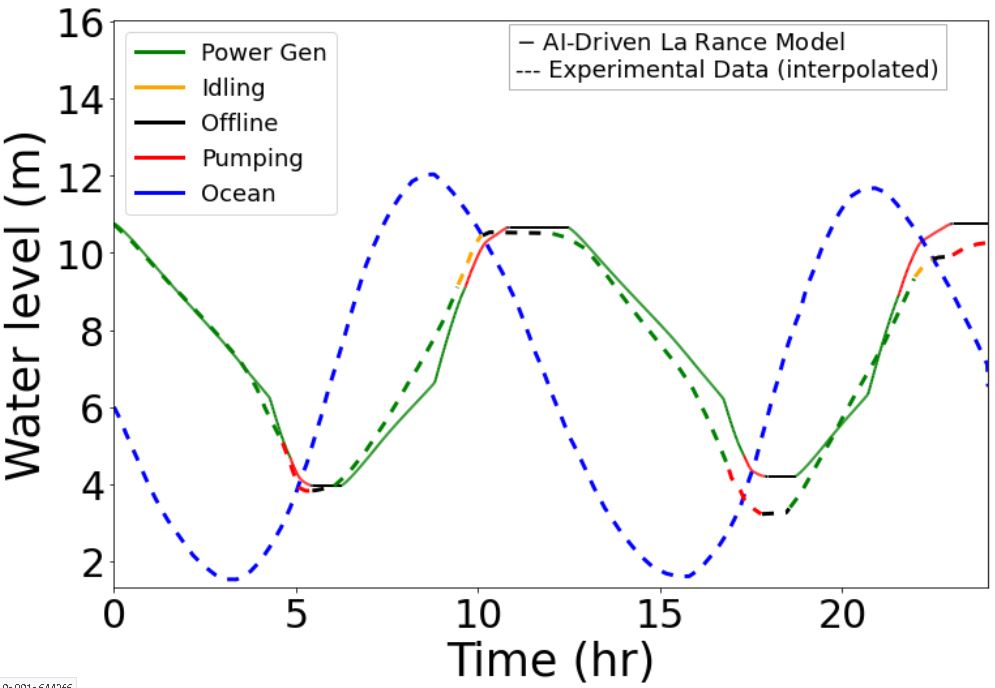}
    \caption{Measured and predicted lagoon water levels for the same real ocean signal.} \label{TWPRwl}
  \end{subfigure} \hfill
  \begin{subfigure}[t]{.45\linewidth}
    \centering\includegraphics[width=\linewidth]{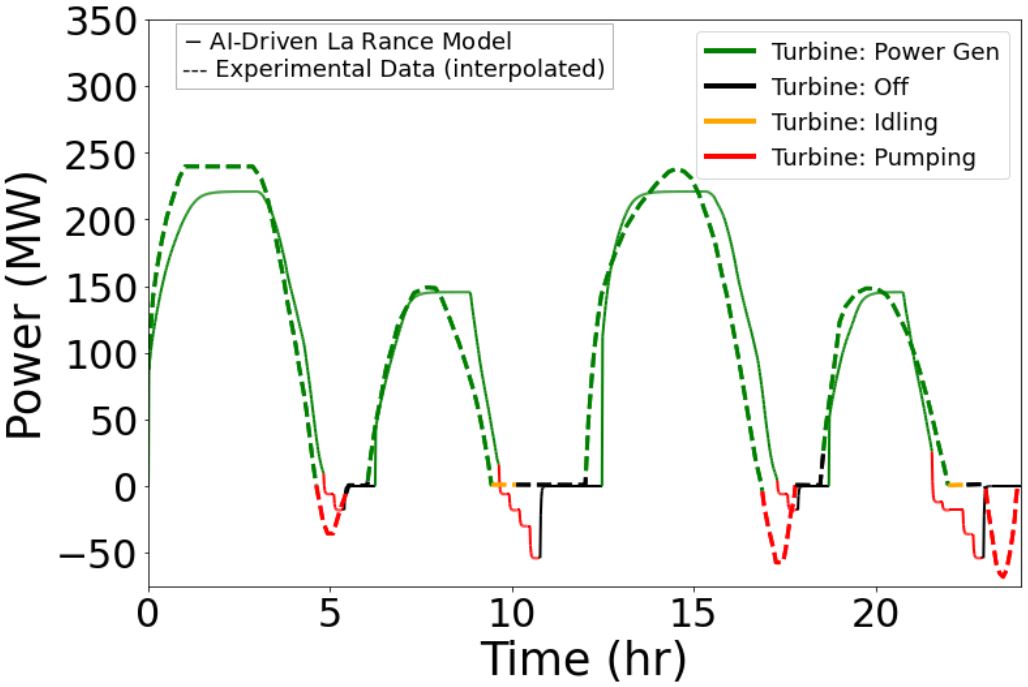}
    \caption{Measured and predicted power output/input for the same real ocean signal.}
    \label{TWPRPow}
  \end{subfigure}
  \caption{Comparisons between measured lagoon water level variations and power output/input from La Rance's $T.W.P$ scheme operation against learned operational strategy by the AI-Driven La Rance model.} \label{LaRancePredvsMeas}
\end{figure}


A comparison of the predicted and measured operation of La Rance, for the same measured tide and starting lagoon water levels from \cite{lebarbier1975power}, is showcased in Fig. \ref{LaRancePredvsMeas}. From the results, we see how the strategy and predicted lagoon water levels of our real-time AI-Driven model closely resembles the results from the $T.W.P$ scheme observed in La Rance, with minor differences in operation probably due to the different goal of optimisation (revenue vs power generation). Together with the verified parametric models for La Rance's hydraulic structures in Section \ref{Val1}, Fig. \ref{LaRancePredvsMeas} completes our model validation against a real TRS.

\section{Conclusion}

In this work, an artificial intelligence (AI) driven representation of a constructed TRS was developed using a 0D TRS parametrised model operated through DRL techniques. Both the 0D TRS model and the AI-Driven representation were then validated, with model predictions showcasing good agreement of results with measured data. For our case study, we utilised the La Rance tidal barrage -- the oldest and most successful TRS ever constructed.

The developed methodologies for reverse engineering the hydraulic structures of La Rance (assembled into the 0D TRS model) are generalisable and can be applied to other constructed TRS. Furthermore, novel representations for lagoon wetted area, turbines in pump mode and momentum ramp functions can be applied to future TRS projects (e.g. Swansea Bay Tidal lagoon).

Once trained with the goal of maximising the net energy output, the DRL agent extracted as much energy from the 0D TRS model as reported by La Rance's measurements for a series of analysed years (with an average gain of 2.6\%). For acquiring such results, the optimal control strategy devised by the trained DRL agent used (i) real-time control of hydraulic structures (ii) fine tuned power input for turbines in pump mode, (iii) pumping with positive head differences (aided by gravity) and (iv) independent operation of sluices. Apart from the latter, these abilities are unique to the developed method (in contrast with constrained state-of-art operation optimisation methods), allowing the AI-Driven representation of La Rance to showcase results which are more realistic than TRS model simulations presented in the current literature.

The broad applicability of the proposed method shall help to develop new projects with more realistic optimised operation of TRS at a moment where renewable and clean sources, such as tidal power, become more attractive to mitigate climate change, while maintaining the possibilities of economic growth at low environmental impact.



\section*{Acknowledgments}
We would like to thank the Brazilian agencies Coordenação de Aperfeiçoamento de Pessoal de Ensino Superior (CAPES) and Conselho Nacional de Desenvolvimento Científico e Tecnológico (CNPq) for providing funding for this research.




\printcredits


\bibliography{cas-refs}





\end{document}